\documentclass[lettersize,journal]{IEEEtran}
\pdfoutput=1
\usepackage{amsmath,amsfonts}
\usepackage{algorithmic}
\usepackage{algorithm}
\usepackage{array}
\usepackage[caption=false,font=normalsize,labelfont=sf,textfont=sf]{subfig}
\usepackage{textcomp}
\usepackage{stfloats}
\usepackage{url}
\usepackage{verbatim}
\usepackage{graphicx}
\usepackage{cite}
\usepackage{booktabs}  
\usepackage{multirow}
\usepackage{xcolor}

\newcommand{\FigureFOne}{
\begin{figure*}[h]
    \centering
    \includegraphics[width=0.79\textwidth]{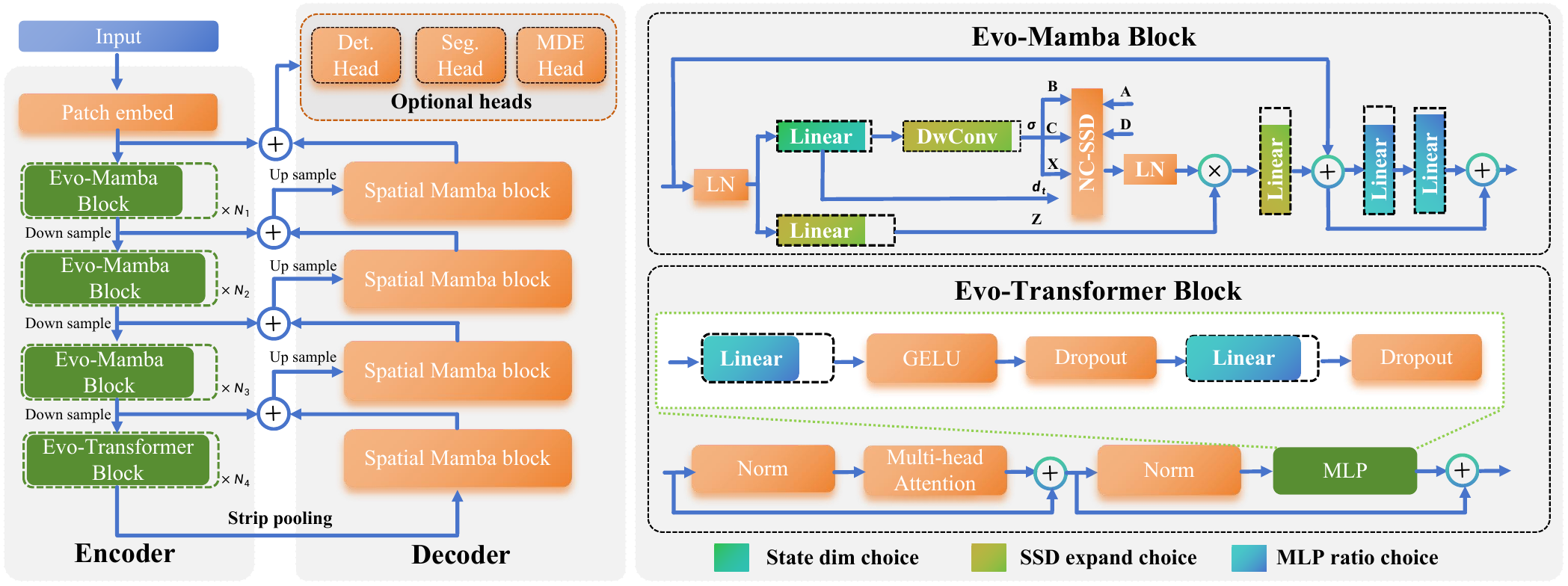}
    \vspace{-10pt}
    \caption{The EvoNAS unified visual architecture, featuring a searchable hybrid VSS-ViT encoder, a Spatial Mamba decoder, and optional task heads for detection, segmentation, and MDE. Searchable components within the Evo-Mamba and Evo-Transformer blocks define the encoder’s supernet search space, enabling flexible and general-purpose architecture optimization across diverse vision tasks.\label{sec:framework}}
    \vspace{-10pt}
\end{figure*}
}

\newcommand{\FigureFTwo}{
\begin{figure}[h]
    \centering
    \includegraphics[width=0.45\textwidth]{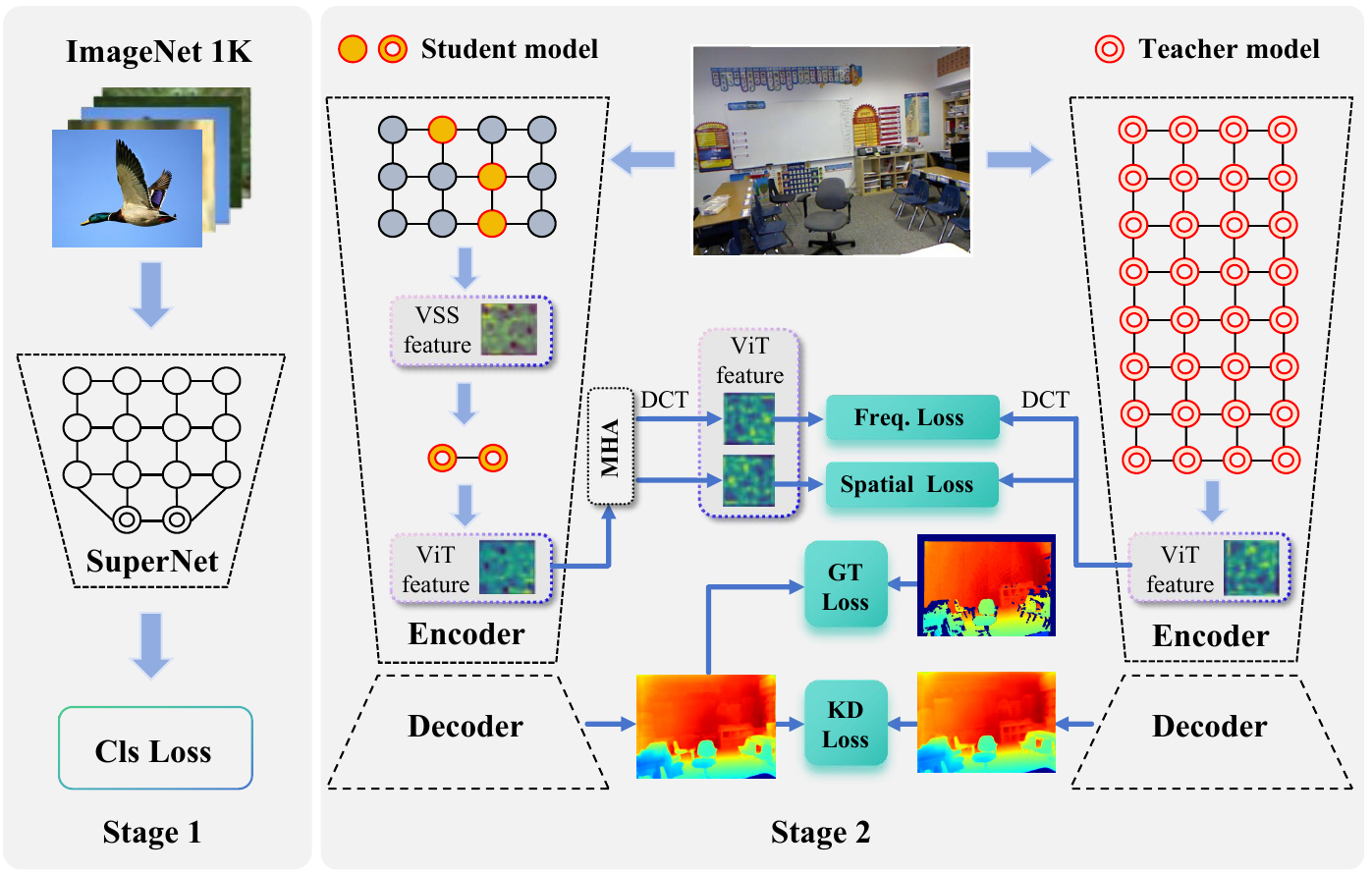}
    \vspace{-10pt}
    \caption{Overview of the two-stage supernet optimization with dual-domain knowledge distillation. Stage~1 pretrains the VSS-ViT supernet encoder on ImageNet-1K to establish a stable initialization. Stage~2 adapts the supernet to the target task by aligning student VSS/ViT features with teacher ViT features through spatial- and frequency-domain distillation, while jointly applying prediction-level and ground-truth supervision.\label{sec:supernet training}}
    \vspace{-10pt}
\end{figure}
}

\newcommand{\FigureFThree}{
\begin{figure*}[ht]
    \centering
    \includegraphics[width=0.93\textwidth]{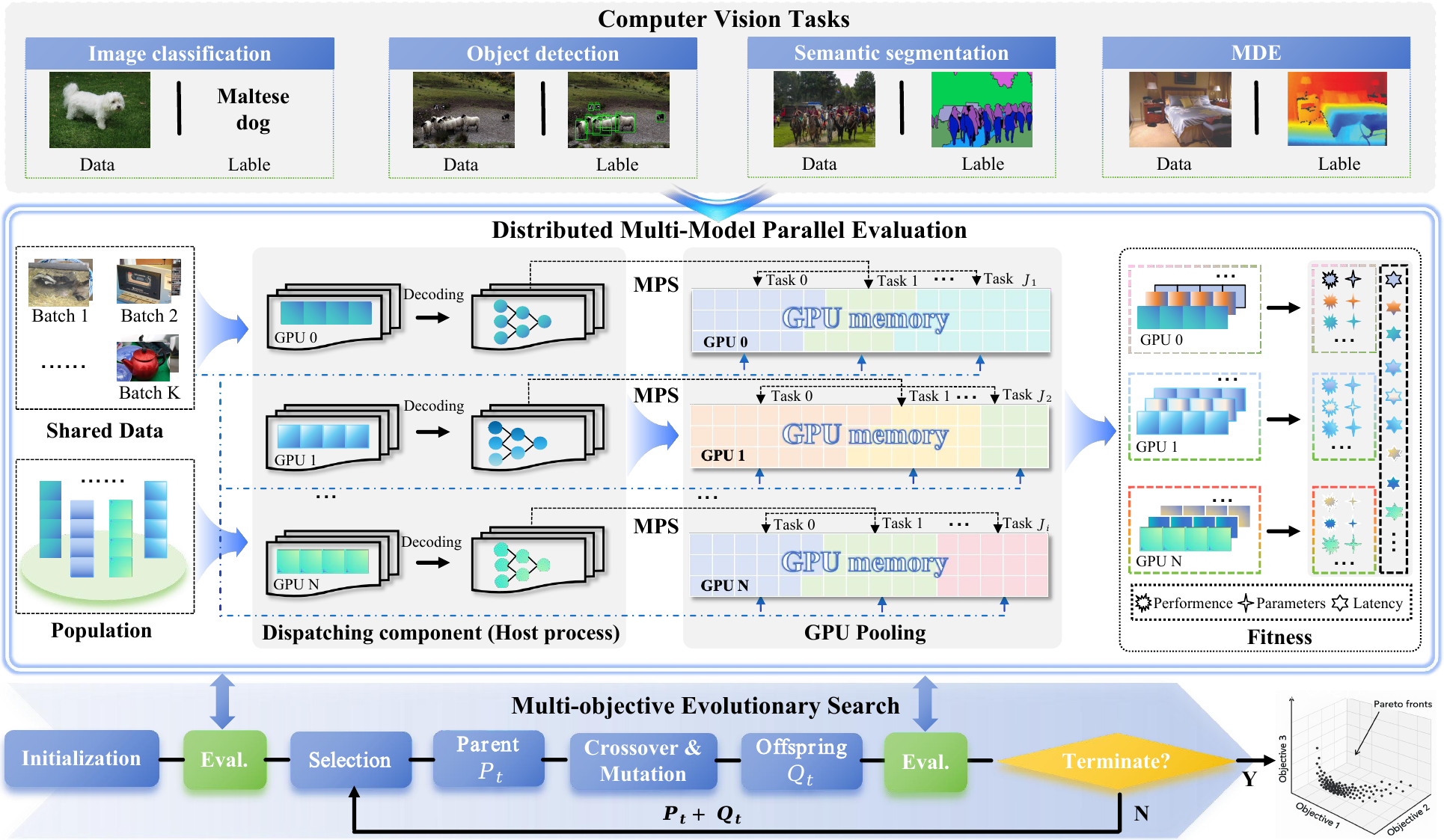}
    \caption{Overview of the EvoNAS pipeline. The framework integrates distributed multi-model parallel evaluation with multi-objective evolutionary search, enabling efficient validation of candidate architectures across GPUs and progressive approximation of the Pareto frontier in terms of accuracy, latency, and computational complexity.}
    \label{sec:EvoX}
\end{figure*}
}

\newcommand{\FigureFFour}{
\begin{figure}[ht]
    \centering
    \includegraphics[width=0.45\textwidth]{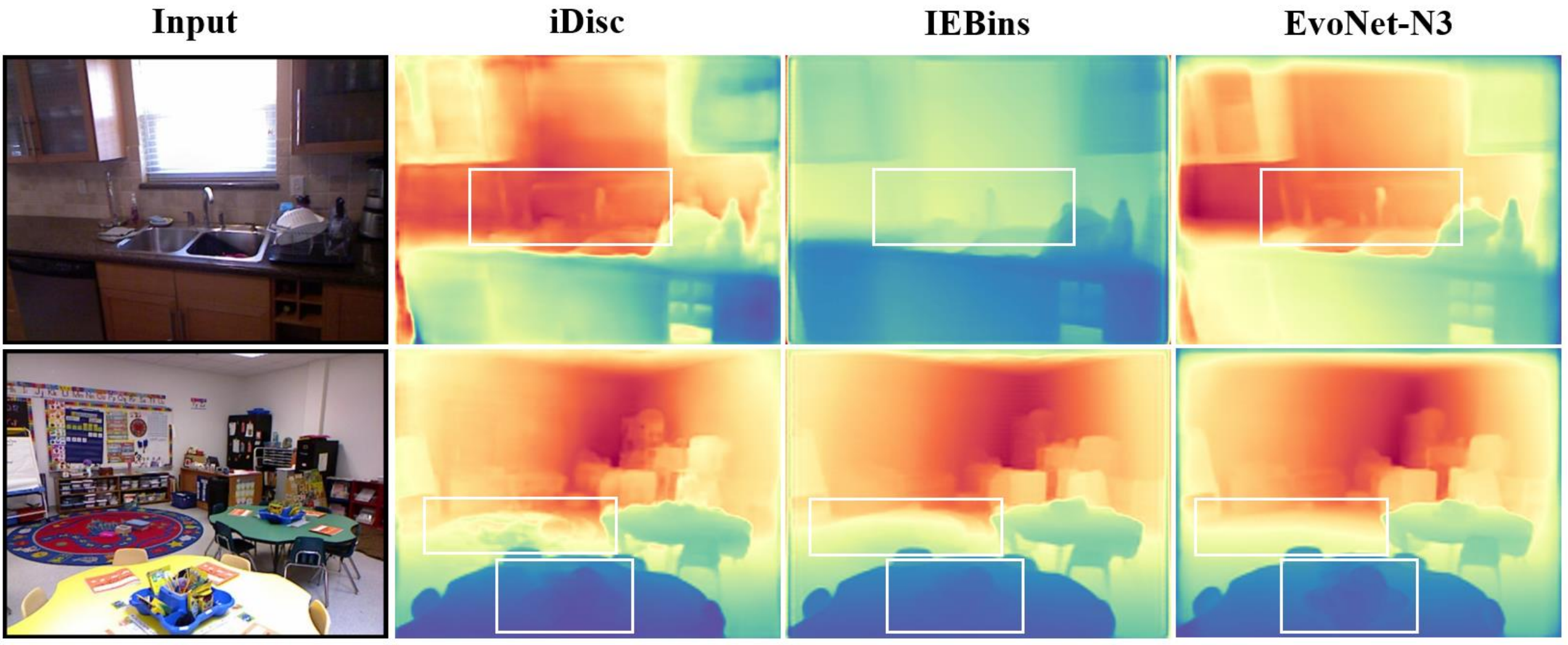}
    \caption{Qualitative comparisons of depth estimation results on NYU dataset. }
    \label{fig:nyu}
\end{figure}
}

\newcommand{\FigureFive}{
\begin{figure}[ht]
    \centering
    \includegraphics[width=0.45\textwidth]{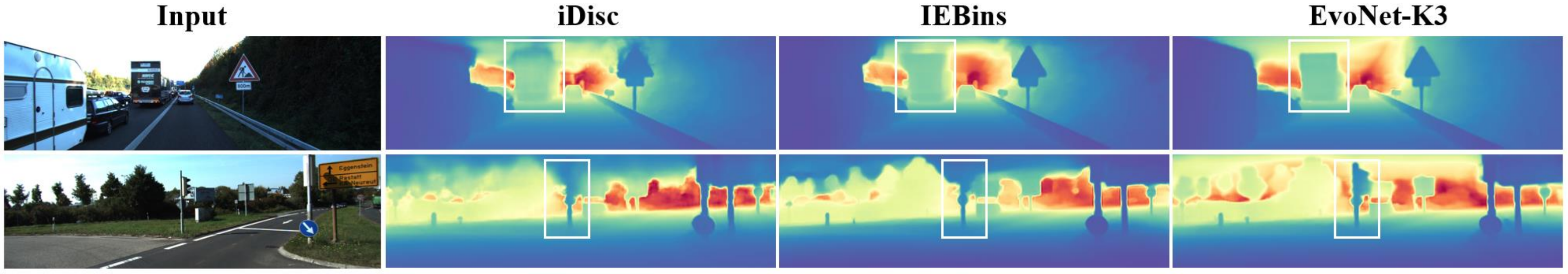}
    \caption{Qualitative comparisons of depth estimation results on KITTI dataset. }
    \label{fig:kitti}
\end{figure}
}

\newcommand{\FigureSix}{
\begin{figure}[ht]
    \centering
    \includegraphics[width=0.45\textwidth]{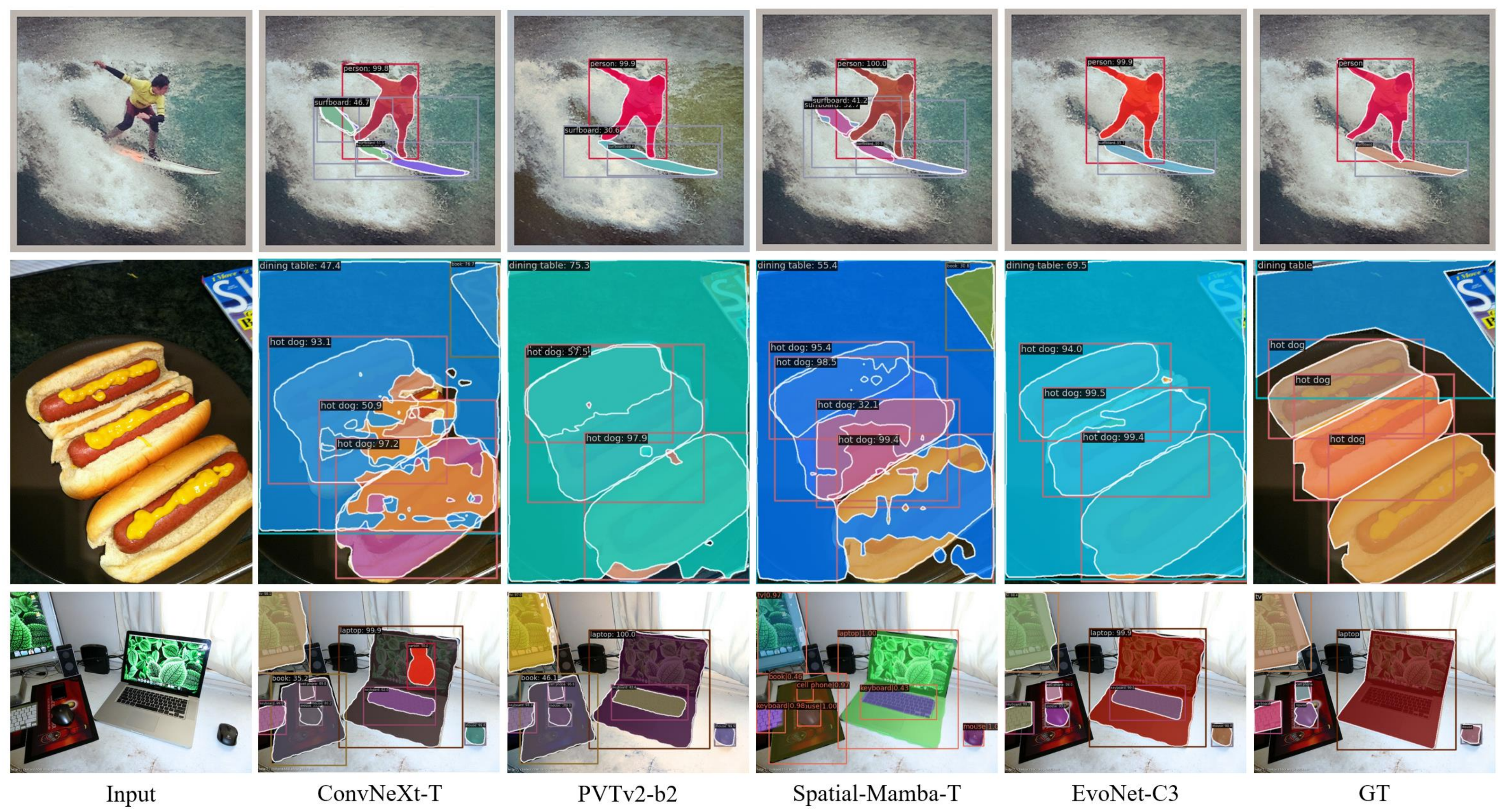}
    \caption{Qualitative comparisons of results on COCO dataset. }
    \label{fig:COCO}
\end{figure}
}

\newcommand{\FigureSeven}{
\begin{figure}[ht]
    \centering
    \includegraphics[width=0.45\textwidth]{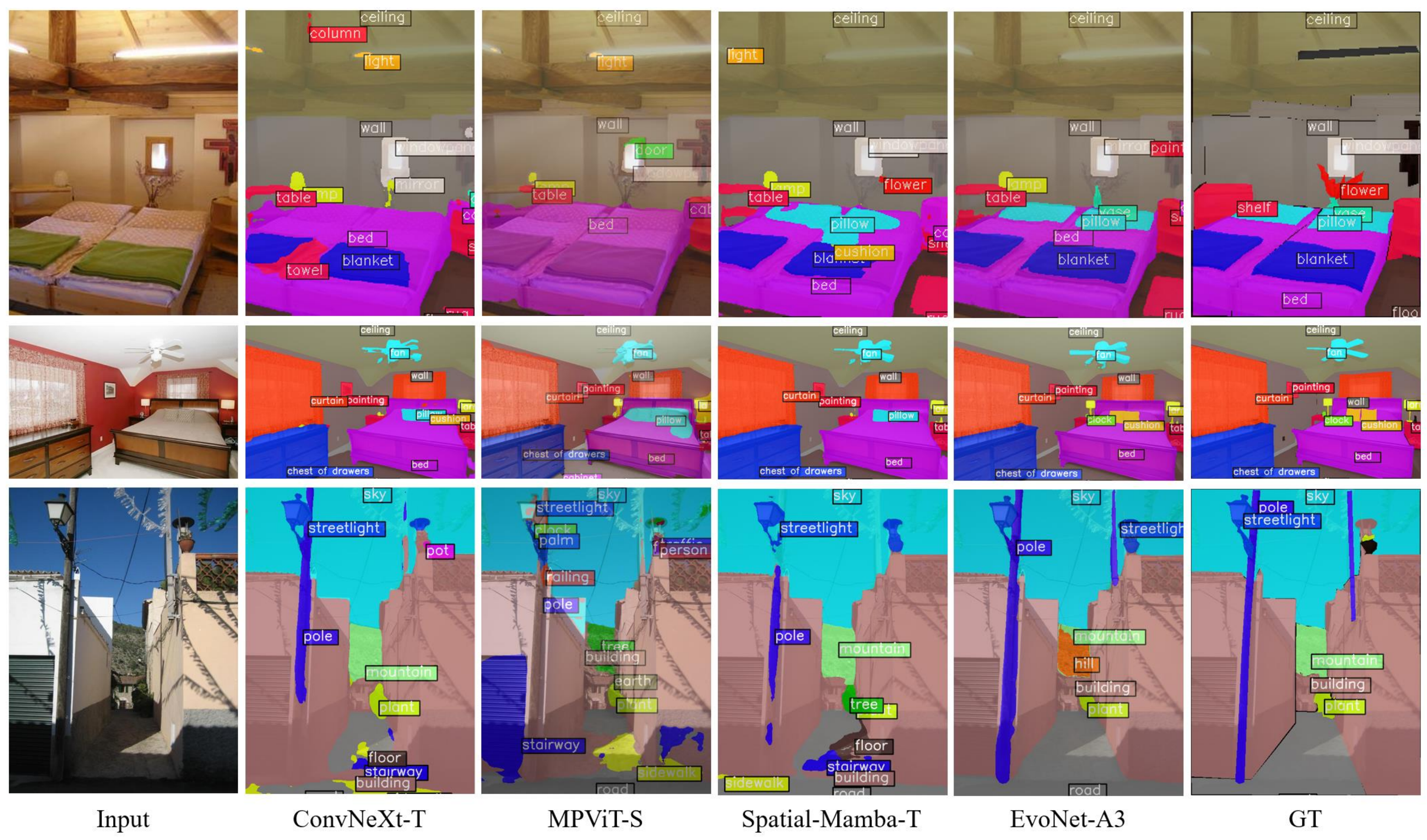}
    \caption{Qualitative comparisons of results on ADE20K dataset. }
    \label{fig:ADE20K}
\end{figure}
}

\newcommand{\Figureeight}{
\begin{figure}[ht]
    \centering
    \includegraphics[width=0.45\textwidth]{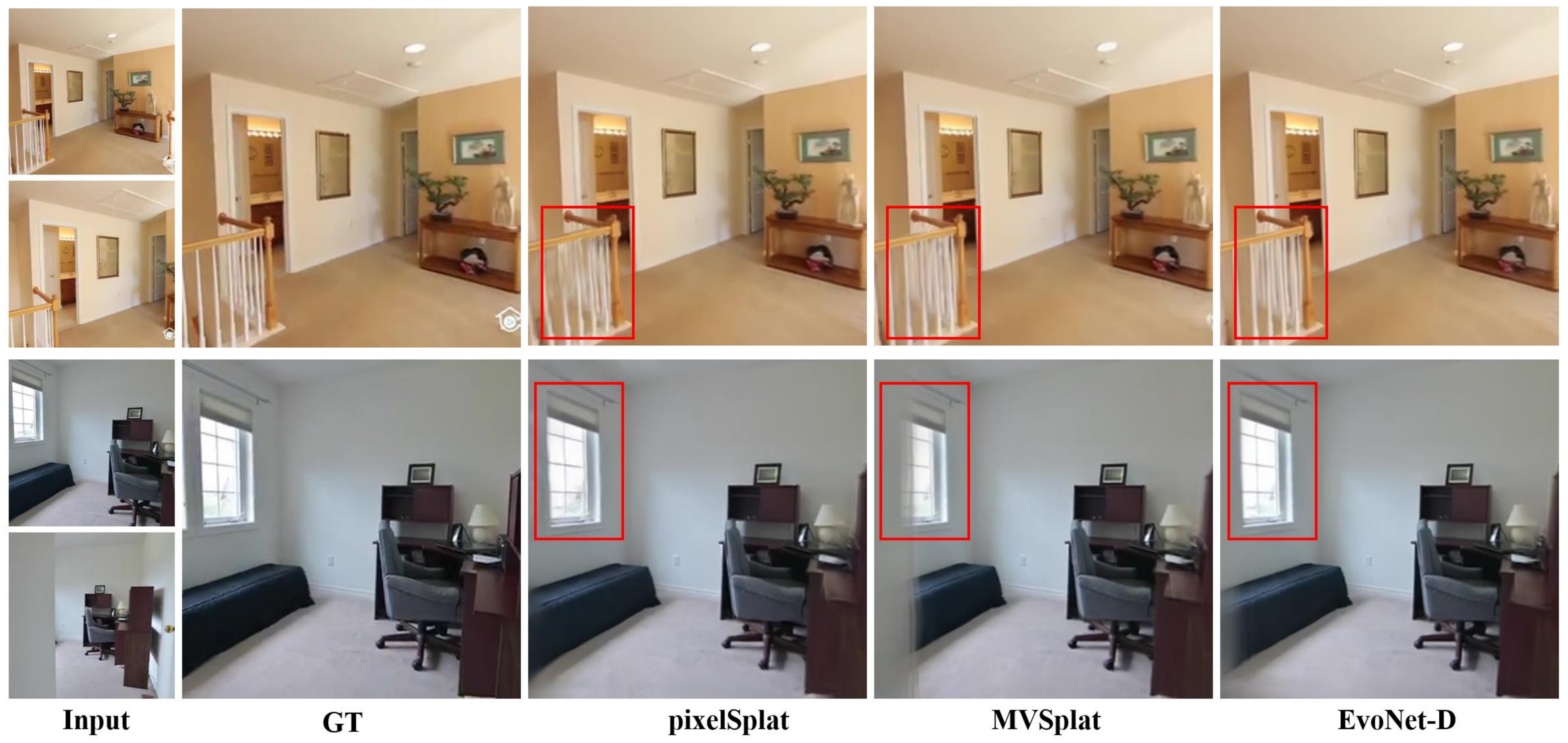}
    \caption{Qualitative comparisons of results on RealEstate10K dataset.}
    \label{fig:NVS}
\end{figure}
}

\newcommand{\Figurenin}{
\begin{figure}[ht]
    \centering
    \includegraphics[width=0.45\textwidth]{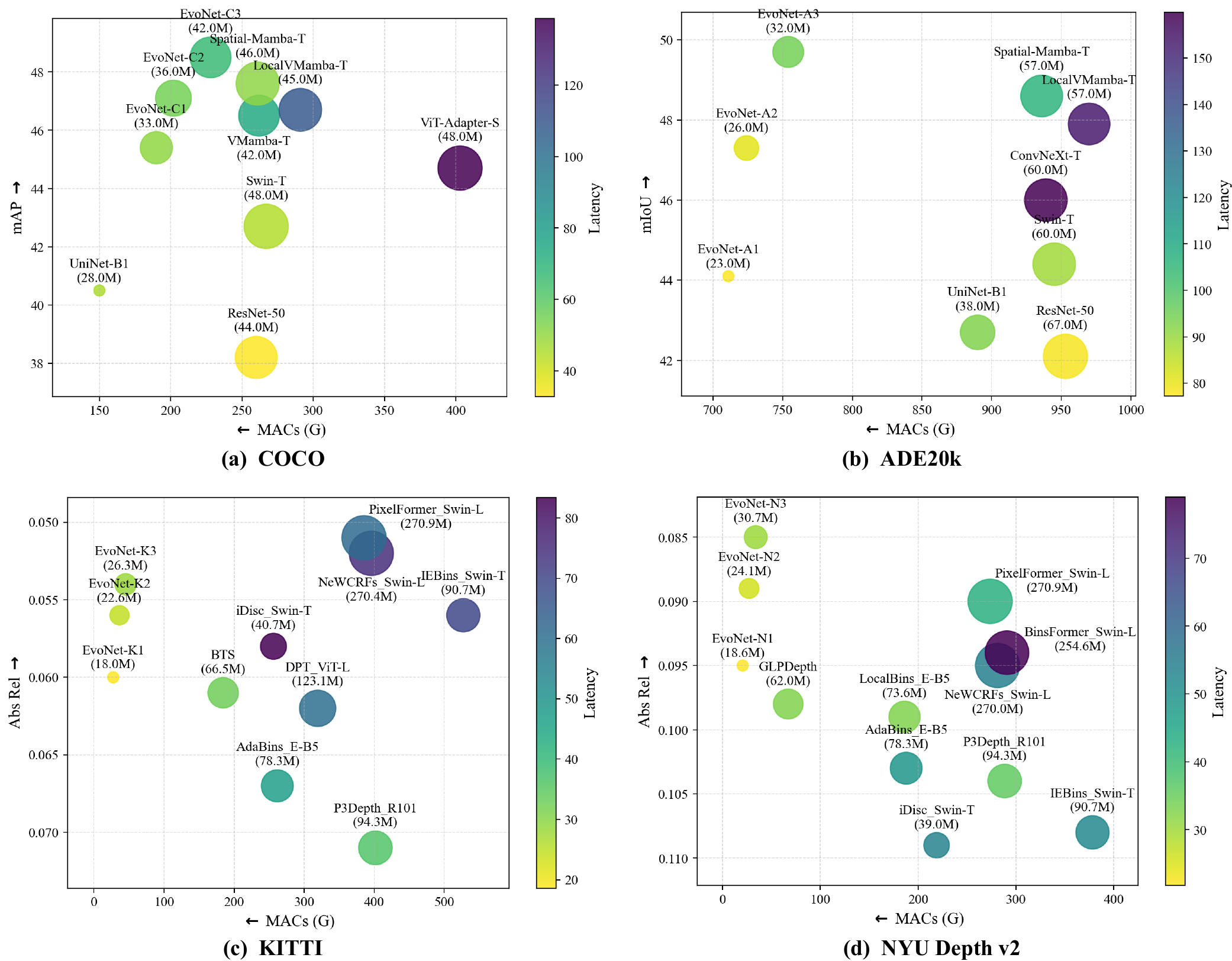}
    \caption{Performance–efficiency trade-offs of EvoNAS across four benchmarks: (a) COCO, (b) ADE20K, (c) KITTI, and (d) NYU v2. The x-axis denotes MACs, the y-axis represents task-specific accuracy metrics (mAP, mIoU, or AbsRel), bubble size indicates parameter count, and color encodes inference latency.}
    \label{fig:overall per}
\end{figure}
}

\newcommand{\Figureten}{
\begin{figure*}[ht]
    \centering
    \includegraphics[width=1.0\textwidth]{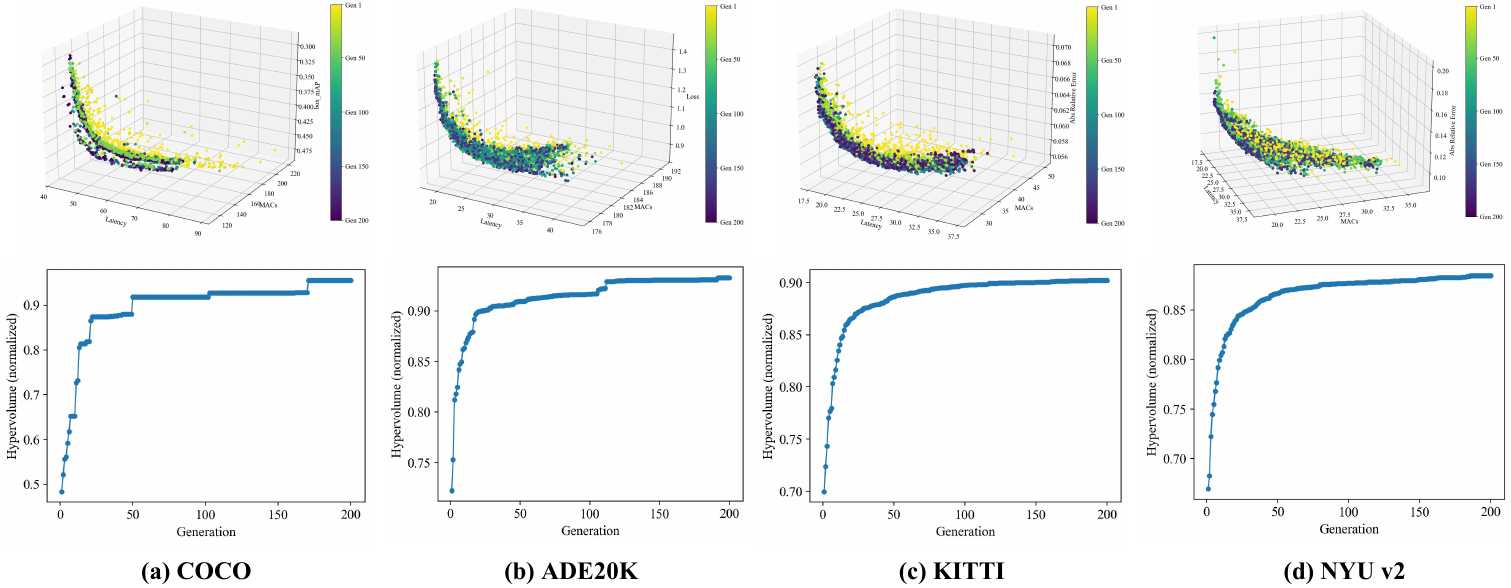}
    \caption{Evolutionary trajectories of EvoNAS on COCO, ADE20K, KITTI, and NYU v2. Top: 3D distribution of candidate architectures across generations in the accuracy–latency–MACs space. Bottom: normalized hypervolume curves showing the convergence behavior of the Pareto front over 200 generations.}
    \label{fig:tra}
\end{figure*}
}

\newcommand{\Figureele}{
\begin{figure}[ht]
    \centering
    \includegraphics[width=0.3\textwidth]{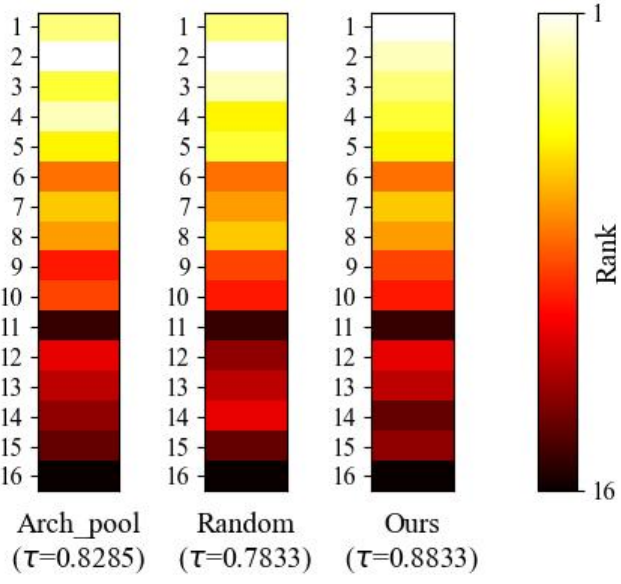}
    \caption{Comparison of architecture ranking consistency.}
    \label{fig:pai}
\end{figure}
}

\providecommand{\TabCOCOResults}{
\begin{table*}[t]
\centering
\caption{Comparison of quantitative results on COCO}
\label{tab:kitti_results}

\end{tabular}
\end{table*}}%

\providecommand{\TabCOCOResult}{
\begin{table}[t]
\centering
\caption{Object detection on COCO with Mask R-CNN~\cite{he2017mask}. Latency and throughput are measured at a test resolution of $800\times1280$.
($\uparrow$ indicates higher is better, $\downarrow$ indicates lower is better.)}
\label{tab:coco_result}
\setlength{\tabcolsep}{3.5pt}
\begin{tabular}{lcccccc}
\toprule
Model & Params & MACs & $\mathrm{AP}^{b} \uparrow$ & Lat.$\downarrow$ & Thrp.$\uparrow$ & NID$\uparrow$ \\
\midrule
ResNet-50~\cite{he2016deep}             & 44M & 260G & 38.2 & 32.8ms  & 34 & 0.87 \\
ConvNeXt-T~\cite{liu2022convnet}        & 48M & 262G & 44.2 & 43.4ms  & 26 & 0.92 \\
Swin-T~\cite{liu2021swin}               & 48M & 267G & 42.7 & 45.1ms  & 24 & 0.89 \\
ViT-Adapter-S~\cite{chenvision}         & 48M & 403G & 44.7 & 138.7ms & 7  & 0.93 \\
MPViT-S~\cite{lee2022mpvit}             & 43M & 268G & 46.4 & --      & -- & 1.08 \\
PVTv2-B2~\cite{wang2022pvt}             & 45M & --   & 45.3 & 86.7ms  & 7  & 1.01 \\
VMamba-T~\cite{liu2024vmamba}           & 42M & 262G & 46.5 & 74.3ms  & 12 & 1.11 \\
LocalVMamba-T~\cite{huang2024localmamba} & 45M & 291G & 46.7 & 108.8ms & 8  & 1.04 \\
Spatial-Mamba-T~\cite{xiao2025spatial}  & 46M & 261G & 47.6 & 51.0ms  & 20 & 1.03 \\
ViTAS-Twins-S~\cite{su2022vitas}        & 44M & 168G & 45.9 & --     & -- & 1.04 \\
UniNet-B1~\cite{liu2022uninet}          & 28M & 150G & 40.5 & 45.7ms & 24 & 1.45 \\
PVT-S+AttnZero~\cite{li2024attnzero}    & --  & 248G & 41.0 & 51.3ms & 23 & - \\
\midrule
\textbf{EvoNet-C1} & \textbf{33M} & \textbf{190G} & \textbf{45.4} & \textbf{50.2ms} & \textbf{26} & \textbf{1.39} \\
\textbf{EvoNet-C2} & \textbf{36M} & \textbf{202G} & \textbf{47.1} & \textbf{55.4ms} & \textbf{23} & \textbf{1.29} \\
\textbf{EvoNet-C3} & \textbf{42M} & \textbf{228G} & \textbf{48.5} & \textbf{66.9ms} & \textbf{18} & \textbf{1.15} \\
\bottomrule
\end{tabular}
\end{table}
}

\providecommand{\TabADEResults}{
\begin{table}[t]
\centering
\caption{Semantic segmentation on ADE20K with UPerNet~\cite{xiao2018unified}. Latency and throughput are measured at a test resolution of $512\times2048$.
($\uparrow$ indicates higher is better, $\downarrow$ indicates lower is better.)}
\label{tab:ade20k_result}
\setlength{\tabcolsep}{3.5pt}
\begin{tabular}{lcccccc}
\toprule
Model & Params & MACs & mIoU$\uparrow$ & Lat.$\downarrow$ & Thrp.$\uparrow$ & NID$\uparrow$ \\
\midrule
ResNet-50~\cite{he2016deep}              & 67M & 953G & 42.1 & 78.3ms  & 13 & 0.63 \\
ConvNeXt-T~\cite{liu2022convnet}         & 60M & 939G & 46.0 & 159.8ms & 6  & 0.77 \\
Swin-T~\cite{liu2021swin}                & 60M & 945G & 44.4 & 89.0ms  & 12 & 0.74 \\
MPViT-S~\cite{lee2022mpvit}              & 52M & 943G & 48.3 & 153.2ms & 9  & 0.93 \\
ViT-Adapter-T~\cite{chenvision}          & 36M & --   & 42.6 & 238.2ms & 6  & 1.18 \\
VMamba-T~\cite{liu2024vmamba}            & 62M & 949G & 48.0 & 118.5ms & 8  & 0.77 \\
LocalVMamba-T~\cite{huang2024localmamba} & 57M & 970G & 47.9 & 153.7ms & 6  & 0.84 \\
Spatial-Mamba-T~\cite{xiao2025spatial}   & 57M & 936G & 48.6 & 106.2ms & 10 & 0.85 \\
UniNet-B1~\cite{liu2022uninet}           & 38M & 890G & 42.7 & 92.9ms  & 10 & 1.12 \\
ViTAS-Twins-S~\cite{su2022vitas}         & 62M & --   & 47.9 & --      & -- & 0.78 \\
S3-T~\cite{chen2021searching}            & 60M & 954G & 44.9 & --      & -- & 0.75 \\
\midrule
\textbf{EvoNet-A1} & \textbf{23M} & \textbf{711G} & \textbf{44.1} & \textbf{77.3ms} & \textbf{14} & \textbf{1.93} \\
\textbf{EvoNet-A2} & \textbf{26M} & \textbf{724G} & \textbf{47.3} & \textbf{81.0ms} & \textbf{13} & \textbf{1.79} \\
\textbf{EvoNet-A3} & \textbf{32M} & \textbf{754G} & \textbf{49.7} & \textbf{94.8ms} & \textbf{12} & \textbf{1.57} \\
\bottomrule
\end{tabular}
\end{table}
}

\providecommand{\TabKittiResults}{
\begin{table}[t]
\centering
\scriptsize
\setlength{\tabcolsep}{3.5pt}
\renewcommand{\arraystretch}{0.95}
\caption{Comparison of quantitative results on KITTI Eigen-Split.
($\uparrow$ indicates higher is better, $\downarrow$ indicates lower is better.)}
\label{tab:kitti_results}
\begin{tabular}{lrrrrrrr}
\toprule
Model & Params & MACs & Lat.$\downarrow$ & Thrp.$\uparrow$ & Abs Rel$\downarrow$ & $\delta_1 \uparrow$ & NID$\uparrow$ \\
\midrule
BTS~\cite{lee2019big}               & 66.5M  & 184.3G & 32.9ms & 35  & 0.061 & 0.954 & 1.43 \\
AdaBins~\cite{bhat2021adabins}      & 78.3M  & 261.7G & 47.2ms & 21  & 0.067 & 0.949 & 1.21 \\
P3Depth~\cite{patil2022p3depth}     & 94.3M  & 401.8G & 36.1ms & 34  & 0.071 & 0.953 & 1.01 \\
DPT~\cite{ranftl2021vision}         & 123.1M & 319.3G & 60.1ms & 19  & 0.062 & 0.959 & 0.78 \\
iDisc~\cite{piccinelli2023idisc}    & 40.7M  & 256.2G & 83.4ms & 14  & 0.058 & 0.968 & 2.38 \\
IEBins~\cite{shao2023iebins}        & 90.7M  & 527.3G & 69.5ms & 16  & 0.056 & 0.970 & 1.07 \\
BinsFormer~\cite{li2024binsformer}  & 254.6M & 409.4G & 97.5ms & 13  & 0.052 & 0.974 & 0.38 \\
NeWCRFs~\cite{yuan2022neural}       & 270.4M & 395.9G & 74.8ms & 14  & 0.052 & 0.975 & 0.36 \\
PixelFormer~\cite{agarwal2023attention} & 270.9M & 385.4G & 60.9ms & 19  & 0.051 & 0.976 & 0.36 \\
\midrule
\textbf{EvoNet-K1} & \textbf{18.0M} & \textbf{27.3G} & \textbf{18.6ms} & \textbf{117} & \textbf{0.060} & \textbf{0.960} & \textbf{5.34} \\
\textbf{EvoNet-K2} & \textbf{22.6M} & \textbf{36.2G} & \textbf{24.6ms} & \textbf{83}  & \textbf{0.056} & \textbf{0.966} & \textbf{4.28} \\
\textbf{EvoNet-K3} & \textbf{26.3M} & \textbf{45.0G} & \textbf{28.0ms} & \textbf{65}  & \textbf{0.054} & \textbf{0.969} & \textbf{3.68} \\
\bottomrule
\end{tabular}
\end{table}}%

\providecommand{\TabNYUResults}{
\begin{table}[t]
\centering
\scriptsize
\setlength{\tabcolsep}{3.5pt} % default 6pt; shrink column spacing
\renewcommand{\arraystretch}{0.95} % shrink row spacing
\caption{Comparison of quantitative results on NYU Depth v2 dataset.
($\uparrow$ indicates higher is better, $\downarrow$ indicates lower is better.)}
\label{tab:nyu_results}
\begin{tabular}{lrrrrrrr}
\toprule
Model & Params & MACs & Lat.$\downarrow$ & Thrp.$\uparrow$ & Abs Rel$\downarrow$ & $\delta_1 \uparrow$ & NID$\uparrow$ \\
\midrule
BTS~\cite{lee2019big}                & 66.5M  & 132.3G & 28.2ms & 45  & 0.113 & 0.871 & 1.31 \\
LocalBins~\cite{bhat2022localbins}   & 73.6M  & 186.2G & 32.9ms & 41  & 0.099 & 0.907 & 1.23 \\
GLPDepth~\cite{kim2022global}        & 62.0M  & 67.2G  & 32.6ms & 52  & 0.098 & 0.915 & 1.48 \\
AdaBins~\cite{bhat2021adabins}       & 78.3M  & 187.8G & 49.2ms & 20  & 0.103 & 0.903 & 1.15 \\
P3Depth~\cite{patil2022p3depth}      & 94.3M  & 288.3G & 35.8ms & 35  & 0.104 & 0.904 & 0.96 \\
iDisc~\cite{piccinelli2023idisc}     & 39.0M  & 218.8G & 53.4ms & 25  & 0.109 & 0.894 & 2.29 \\
IEBins~\cite{shao2023iebins}         & 90.7M  & 378.2G & 52.0ms & 22  & 0.108 & 0.893 & 0.98 \\
NeWCRFs~\cite{yuan2022neural}        & 270.4M & 281.1G & 54.5ms & 20  & 0.095 & 0.923 & 0.34 \\
PixelFormer~\cite{agarwal2023attention} & 270.9M & 273.5G & 42.8ms & 27  & 0.090 & 0.929 & 0.34 \\
BinsFormer~\cite{li2024binsformer}   & 254.6M & 290.7G & 79.0ms & 18  & 0.094 & 0.925 & 0.36 \\
\midrule
\textbf{EvoNet-N1} & \textbf{19.1M} & \textbf{21.7G} & \textbf{21.8ms} & \textbf{138} & \textbf{0.095} & \textbf{0.912} & \textbf{4.77} \\
\textbf{EvoNet-N2} & \textbf{24.1M} & \textbf{27.1G} & \textbf{25.9ms} & \textbf{107} & \textbf{0.089} & \textbf{0.926} & \textbf{3.85} \\
\textbf{EvoNet-N3} & \textbf{30.3M} & \textbf{33.9G} & \textbf{30.8ms} & \textbf{88}  & \textbf{0.085} & \textbf{0.932} & \textbf{3.08} \\
\bottomrule
\end{tabular}
\end{table}}%

\providecommand{\TabEvoXNASResults}{
\begin{table}[t]
\centering
\caption{Efficiency validation of the proposed parallel evaluation framework. PWs denotes \textit{Persistent Workers}. \textit{Config} is reported as (\textit{Proc./GPU}, \textit{Models/Proc.}).}
\label{tab:EvoXNAS}
\setlength{\tabcolsep}{6pt}
\begin{tabular}{llccc}
\toprule
Dataset & Method & Config & $\mathrm{Time(s)}/G$ & Ratio(\%) \\
\midrule
\multirow{6}{*}{NYUv2}
& Baseline                        & --          & 705.1 & 100.0 \\
& Baseline w/ PWs                 & --          & 250.0 & 35.5  \\
& \multirow{4}{*}{DMMPE} 
&                                    (1,4)       & 236.4 & 33.5  \\
&                                  & (4,1)       & 181.7 & 25.8  \\
&                                  & (3,4)       & 178.2 & 25.3  \\
&                                  & (4,3)       & \textbf{172.3} & \textbf{24.4} \\
\midrule
\multirow{6}{*}{KITTI}
& Baseline                       & --          & 756.8 & 100.0 \\
& Baseline + PWs                 & --          & 329.4 & 43.5  \\
& \multirow{4}{*}{DMMPE} 
&                                  (1,4)       & 319.4 & 42.2  \\
&                                  & (4,1)       & 248.9 & 32.9  \\
&                                  & (3,4)       & 243.5 & 32.2  \\
&                                  & (4,3)       & \textbf{231.0} & \textbf{30.5} \\
\bottomrule
\end{tabular}
\end{table}
}

\providecommand{\TabAblationSupernetTraining}{
\begin{table}[t]
\centering
\caption{Ablation analysis of the supernet training strategy.}
\label{tab:AblationSupernetTraining}
\setlength{\tabcolsep}{6pt}
\begin{tabular}{lccc}
\toprule
Method & Abs Rel$\downarrow$ & RMSE$\downarrow$ & $\delta_1\uparrow$ \\
\midrule
\multicolumn{3}{l}{\textit{Target: EvoNet-N1}} \\
Arch. Pool~\cite{zhang2020one}                  & 0.118 & 0.403 & 0.872 \\
Random Sampling                                 & 0.109 & 0.377 & 0.886 \\
\textbf{PST(Ours)}   & \textbf{0.095} & \textbf{0.344} & \textbf{0.912} \\
\midrule
\multicolumn{3}{l}{\textit{Target: EvoNet-N2}} \\
Arch. Pool~\cite{zhang2020one}                  & 0.096 & 0.343 & 0.909 \\
Random Sampling                                 & 0.095 & 0.341 & 0.915 \\
\textbf{PST(Ours)}   & \textbf{0.089} & \textbf{0.323} & \textbf{0.926} \\
\midrule
\multicolumn{3}{l}{\textit{Target: EvoNet-N3}} \\
Arch. Pool~\cite{zhang2020one}                  & 0.093 & 0.332 & 0.920 \\
Random Sampling                                 & 0.090 & 0.321 & 0.925 \\
\textbf{PST(Ours)}   & \textbf{0.085} & \textbf{0.310} & \textbf{0.932} \\
\bottomrule
\end{tabular}
\end{table}
}

\providecommand{\TabAblationDecoderArchitecture}{
\begin{table}[t]
\centering
\caption{Ablation analysis of the decoder architecture.}
\label{tab:AblationDecoderArchitecture}
\setlength{\tabcolsep}{3.5pt}
\begin{tabular}{lccccc}
\toprule
Decoders & MACs & Lat. & Thrp. & Abs Rel & $\delta_1$ \\
\midrule
NewCRFs~\cite{yuan2022neural}    & 32.09G & 37.8ms & 76 & 0.088 & 0.929 \\
iDisc~\cite{piccinelli2023idisc}      & 60.84G & 38.1ms & 53 & 0.091 & 0.922 \\
Spatial-Mamba~\cite{xiao2025spatial} & 33.93G & 30.4ms & 90 & \textbf{0.085} & \textbf{0.932} \\
\bottomrule
\end{tabular}
\end{table}
}

\providecommand{\TabNVSResult}{
\begin{table}[t]
\centering
\caption{NVS on RealEstate10K. Latency and throughput are measured at a test resolution of $256\times256$.}
\label{tab:nvs_result}
\setlength{\tabcolsep}{3.5pt}
\begin{tabular}{lcccccc}
\toprule
Method & PSNR$\uparrow$ & SSIM$\uparrow$ & LPIPS$\downarrow$ & Params & Lat. & Thrp. \\
\midrule
pixelNeRF~\cite{yu2021pixelnerf}         & 20.43 & 0.589 & 0.550 & --  & --     & -- \\
GPNR~\cite{suhail2022generalizable}      & 24.11 & 0.793 & 0.255 & --  & --     & -- \\
AttnRend~\cite{du2023learning}           & 24.78 & 0.820 & 0.213 & --  & --     & -- \\
MuRF~\cite{xu2024murf}                   & 26.10 & 0.858 & 0.143 & --  & --     & -- \\
pixelSplat~\cite{charatan2024pixelsplat} & 25.89 & 0.858 & 0.142 & 119M & 162ms & -- \\
MVSplat~\cite{chen2024mvsplat}           & 26.39 & 0.869 & 0.128 & 12M  & 62ms  & --  \\
DepthSplat~\cite{xu2025depthsplat}       & 27.47 & 0.889 & 0.114 & 354M & 141ms & 13 \\
\textbf{EvoNet-D} & \textbf{26.41} & \textbf{0.871} & \textbf{0.127} & \textbf{44M} & \textbf{88ms} & \textbf{27} \\
\bottomrule
\end{tabular}
\end{table}
}

\providecommand{\TabImageNetResults}{
\begin{table}[t]
\centering
\caption{Image classification results on ImageNet-1K.}
\label{tab:imagenet1k_results}
\setlength{\tabcolsep}{6pt}
\begin{tabular}{lcc}
\toprule
Model & Params & Top-1 \\
\midrule
NAT-S~\cite{hassani2023neighborhood} (CVPR'2023)      & 51M & 83.7\% \\
VMamba-B~\cite{liu2024vmamba} (NeurIPS'2024)          & 89M & 83.9\% \\
MILA-T~\cite{han2024demystify} (CVPR'2025)            & 25M & 83.5\% \\
GroupMamba-S~\cite{shaker2025groupmamba} (CVPR'2025)  & 34M & 83.9\% \\
\textbf{EvoNet}                         & \textbf{17M} & \textbf{84.3\%} \\
\bottomrule
\end{tabular}
\end{table}
}

\providecommand{\TabKnowledgeDistillation}{
\begin{table}[t]
\centering
\caption{Fusion Experiment of Knowledge Distillation Strategy}
\label{tab:knowledge_distillation}
\setlength{\tabcolsep}{3.5pt}
\begin{tabular}{lcccccc}
\toprule
Model & Method  & Abs Rel$\downarrow$ & Log10$\downarrow$ & RMSE$\downarrow$ & $\delta_1 \uparrow$ & $\delta_2 \uparrow$ \\
\midrule
\multirow{2}{*}{EvoNet-N1} & w/o CA-DDKD & 0.102 & 0.044 & 0.367 & 0.899 & 0.982 \\
                        & -     & \textbf{0.095} & \textbf{0.041} & \textbf{0.344} & \textbf{0.912} & \textbf{0.985} \\
\multirow{2}{*}{EvoNet-N2} & w/o CA-DDKD & 0.093 & 0.040 & 0.332 & 0.919 & 0.988 \\
                        & -     & \textbf{0.089} & \textbf{0.038} & \textbf{0.323} & \textbf{0.926} & \textbf{0.988} \\
\multirow{2}{*}{EvoNet-N3} & w/o CA-DDKD & 0.090 & 0.039 & 0.328 & 0.922 & 0.989 \\
                        & -     & \textbf{0.085} & \textbf{0.037} & \textbf{0.310} & \textbf{0.932} & \textbf{0.990} \\
\bottomrule
\end{tabular}
\end{table}
}

\begin{document}

\title{Dual-Domain Representation Alignment: Bridging 2D and 3D Vision via Geometry-Aware Architecture Search}

\author{IEEE Publication Technology,~\IEEEmembership{Staff,~IEEE,}

\author{Haoyu~Zhang, Zhihao~Yu, Rui~Wang, Yaochu~Jin,~\ Qiqi~Liu,~\ and~Ran~Cheng,~\

\thanks{Haoyu Zhang, Zhihao Yu, Rui~Wang, are with Hangzhou Normal University, Hangzhou, P.R. China (e-mail: haoyu.zhang@hznu.edu.cn; sakura.yzh391@gmail.com; kujiyunmili@gmail.com;.)}%
\thanks{Yaochu Jin and Qiqi Liu are with the Trustworthy and General AI Lab, School of Engineering, Westlake University, Hangzhou, P.R. China (e-mail: jinyaochu@westlake.edu.cn; liuqiqi@westlake.edu.cn).}%
\thanks{Ran Cheng is with the Department of Data Science and Artificial Intelligence, and the Department of Computing, The Hong Kong Polytechnic University, Hong Kong SAR, China. He is also with The Hong Kong Polytechnic University Shenzhen Research Institute, Shenzhen, China (E-mail: ranchengcn@gmail.com)}%
}

\thanks{This paper was produced by the IEEE Publication Technology Group. They are in Piscataway, NJ.}
\thanks{Manuscript received April 19, 2021; revised August 16, 2021.}}

\markboth{Journal of \LaTeX\ Class Files,~Vol.~14, No.~8, August~2021}
{Shell \MakeLowercase{\textit{et al.}}: A Sample Article Using IEEEtran.cls for IEEE Journals}

\maketitle

\begin{abstract}
Modern computer vision tasks require a balance between predictive accuracy and real-time efficiency. However, the substantial inference cost of large vision models (LVMs) notably restricts their deployment on resource-constrained edge devices. Although Evolutionary Neural Architecture Search (ENAS) is suitable for multi-objective optimization, its practical effectiveness is limited by two challenges: excessive candidate evaluation overhead and significant ranking inconsistency among subnetworks. To address these issues, we propose EvoNAS, a framework for multi-objective evolutionary architecture search that is both efficient and distributed. We first construct a hybrid supernet that integrates Vision State Space and Vision Transformer (VSS-ViT) modules, and we optimize it using a novel Cross-Architecture Dual-Domain Knowledge Distillation (CA-DDKD) strategy. By facilitating interaction between the computational efficiency of VSS blocks and the semantic representation of ViT modules, CA-DDKD substantially enhances the representational capacity of the shared supernet. This strategy also improves the ranking consistency among candidate subnetworks, thereby enabling reliable fitness estimation during evolution without additional fine-tuning. To alleviate the computational bottleneck of large-scale validation, we develop a Distributed Multi-Model Parallel Evaluation (DMMPE) framework based on GPU resource pooling and asynchronous scheduling. Compared with conventional data-parallel evaluation, DMMPE improves evaluation efficiency by over 70\% through concurrent multi-GPU and multi-model execution. Experiments on vision benchmarks, including COCO, ADE20K, KITTI, and NYU-Depth v2, demonstrate that the searched architectures, called EvoNets, consistently establish Pareto-optimal trade-offs between accuracy and efficiency. Compared with representative CNN-, ViT-, and Mamba-based models, EvoNets achieve lower inference latency and higher throughput under strict computational budgets, while they maintain strong generalization capability on downstream tasks such as novel view synthesis. Code is available at https://github.com/EMI-Group/evonas
\end{abstract}

\section{Introduction}

\IEEEPARstart{T}{he} evolution of modern computer vision requires an architectural paradigm that is unified and capable of bridging the gap between dense pixel prediction in 2D and high-fidelity rendering in 3D through the extraction of robust geometric priors. Tasks such as object detection, semantic segmentation, and monocular depth estimation (MDE) necessitate models that process high-resolution inputs and produce outputs that are both dense and spatially structured under strict latency constraints. Recent advances have been driven by the emergence of LVMs and foundation architectures, which establish state-of-the-art performance across a broad spectrum of vision tasks. Representative examples include transformer-based backbones \cite{liu2021swin, oquab2023dinov2}, structured state space models \cite{liu2024vmamba}, and pre-trained models that are domain-specific, such as the Depth Anything series \cite{yang2024depth, lin2025depth}. While these models demonstrate strong generalization by modeling global context, they often lack the explicit geometric grounding required for sensitive 3D tasks, such as high-fidelity neural rendering.

Large vision models exhibit inherent limitations that restrict their practical deployment in geometry-sensitive applications. Their substantial parameter scale and high computational complexity result in significant inference overhead, which renders them unsuitable for devices operating under strict latency or energy constraints. Moreover, the scale of these models frequently necessitates centralized cloud deployment, thereby requiring continuous transmission of visual data from edge devices such as autonomous vehicles and mobile systems. This reliance on remote computation introduces additional system-level challenges, including increased communication latency, bandwidth limitations, and privacy risks. Consequently, these factors constrain the applicability of LVMs in real-world scenarios where responsiveness, reliability, and data locality are essential, particularly when the task demands the preservation of geometric boundaries that are fine-grained.

To address these deployment limitations, Neural Architecture Search (NAS) has emerged as an effective paradigm for adapting network architectures to specific hardware constraints and application requirements \cite{jiang2025accelerating}. In resource-constrained scenarios, architectural design must simultaneously consider multiple competing objectives, including predictive accuracy, parameter efficiency, and inference latency. These objectives are inherently conflicting, which formulates the problem as a multi-objective optimization task where improvement in one metric often results in compromise in another \cite{lu2021neural}. EvoNAS provides a framework that is both principled and flexible for this setting by maintaining a population of diverse candidate architectures and progressively selecting non-dominated solutions. This process yields a spectrum of architectures that satisfy varying deployment requirements and computational budgets \cite{liu2021survey}.Conventional weight-sharing NAS frameworks are hindered by representation collapse, where shared weights fail to preserve the high-frequency geometric information required for dense 2D and 3D tasks. This phenomenon results in ranking inconsistency, where the relative performance of subnetworks during the search phase does not correlate with their standalone performance after independent training \cite{zhang2020one}. While various acceleration strategies have been proposed, these methods often neglect the structural integrity of the feature space. For instance, reducing population sizes reduces diversity \cite{lu2020multiobjective}, while surrogate-assisted evaluation remains limited by the approximation errors of proxy models \cite{lu2021neural, sun2019surrogate}. Even one-shot paradigms \cite{pham2018efficient} often fail to maintain the spectral fidelity of features, which leads to geometric blindness where the searched models overlook sharp boundaries and fine textures.

To address this representation collapse, we propose a geometry-aware architecture search paradigm based on a novel CA-DDKD strategy. This strategy is founded on the observation that ranking inconsistency stems from the loss of high-frequency spectral signals during weight sharing. Consequently, CA-DDKD anchors subnetwork features in both spatial and frequency domains using Discrete Cosine Transform (DCT) constraints. By constraining the high-frequency spectral components, the supernet maintains boundary precision regardless of the subnetwork topology. This approach ensures that the fitness estimation of a subnetwork reflects its actual geometric modeling capacity, thereby providing a reliable foundation for multi-objective optimization without the requirement for exhaustive standalone fine-tuning.

The search for these geometry-aware structures is further enabled by a hardware-isolated evaluation engine that purifies the fitness landscape by removing computational noise. While recent advances in Graphics Processing Unit (GPU) acceleration, such as EvoX \cite{huang2024evox}, EvoJAX \cite{tang2022evojax}, and evosax \cite{lange2023evosax}, utilize parallelism to increase throughput, these systems often encounter latency jitter. Inference latency is highly sensitive to the runtime state of the GPU, which includes kernel launch patterns and resource contention. When multiple architectures are executed concurrently, interference among CUDA kernels can distort timing behavior and render latency measurements inconsistent. To resolve this issue, we develop a DMMPE engine that utilizes hardware isolation to guarantee unbiased physical latency measurements, which ensures the evolutionary algorithm is guided by physical truth rather than scheduling artifacts.

The resulting architectures, designated as EvoNets, establish Pareto-optimal benchmarks in 2D and 2.5D vision and demonstrate significant transferability to 3D Gaussian Splatting (3DGS). Extensive experiments on object detection, semantic segmentation, and MDE confirm the robust performance of the discovered encoders. Specifically, when applied to novel view synthesis on the RealEstate10K benchmark, the proposed geometry-aware encoder achieves high rendering fidelity and reduces the parameter overhead by 88\% (44M vs. 354M) compared to existing baselines. This performance provides empirical evidence that the proposed framework captures universal geometric priors rather than overfitting to specific 2D patterns, thereby validating the efficacy of dual-domain alignment.

The main contributions of this paper are summarized as follows:

\begin{itemize}\item We develop a DMMPE framework based on GPU resource pooling, which integrates GPU virtualization with asynchronous task scheduling to eliminate computational noise. This design ensures unbiased physical latency measurements and high-throughput evaluation, thereby providing a ``clean-room'' environment for evolutionary optimization.

\item We propose a novel supernet optimization strategy driven by CA-DDKD. By enforcing alignment in both spatial and frequency domains via DCT, CA-DDKD mitigates representation collapse and substantially improves ranking consistency, thereby enabling reliable fitness estimation across heterogeneous subnetworks.

\item We design a hybrid VSS-ViT search space that integrates the linear-time computational efficiency of Vision State Space (VSS) blocks with the global reasoning of Vision Transformers (ViT). This hybrid manifold establishes an expressive foundation for capturing universal geometric priors, which enables seamless generalization from 2D pixel prediction to 3DGS.

\end{itemize}

The remainder of this paper is organized as follows. Section~\ref{sec:related_work} reviews the related literature regarding NAS and geometric vision. Section~\ref{sec:architecture} presents the unified visual architecture in detail. Section~\ref{sec:NAS} introduces the EvoNAS framework and the dual-domain optimization procedure. Experimental settings and results are reported in Section~\ref{sec:experiments} and Section~\ref{sec:experiments_ss}. Section~\ref{sec:application} investigates the application of EvoNAS to 3DGS. Finally, Section~\ref{sec:conclusion} concludes the paper and discusses future directions for geometry-aware architecture evolution.

\section{Related Work}
\label{sec:related_work}

While early architectural innovations focused on global semantic extraction, dense prediction tasks necessitate a shift toward the preservation of spatial hierarchies. Consequently, the transition to 3D rendering requires backbones that maintain geometric fidelity across multiple scales, which ensures high-resolution structural accuracy.

NAS enables the efficient discovery of task-specific topologies, yet weight-sharing paradigms frequently suffer from representation collapse and ranking inconsistency. To address this, structural alignment in both spatial and frequency domains is required, which stabilizes the feature-modeling capabilities of heterogeneous subnetworks.

Evolutionary Multi-Objective Optimization (EMO) provides a robust mechanism for navigating the accuracy-latency Pareto front, although its efficacy is sensitive to the precision of the underlying fitness landscape. However, hardware-aware evaluation often introduces latency jitter, which necessitates the use of distributed engines that utilize hardware isolation to ensure measurement reliability.

\subsection{NAS for computer vision tasks}The evolution of NAS has progressed from the optimization of coarse-grained semantic classifiers toward the discovery of structural hierarchies that preserve complex spatial features. Early NAS research focused primarily on image classification, where reinforcement learning (RL) and evolutionary computation (EC) were employed to explore cell-level building blocks or full-network topologies on benchmarks such as CIFAR-10 and ImageNet. Representative works, including NASNet \cite{zoph2018learning} and AmoebaNet \cite{real2019regularized}, demonstrated the potential of automated design by achieving performance that outperformed manual engineering. However, these early approaches incurred substantial computational costs because they required the training of each candidate architecture from scratch. Consequently, the resulting search procedures often required thousands of GPU hours, thereby limiting the scalability and practical applicability of early NAS frameworks. To address this efficiency bottleneck, the field transitioned toward weight-sharing paradigms that enable rapid architectural evaluation.

The expansion of NAS into dense prediction necessitates geometry-aware backbones that maintain high-resolution feature maps and multi-scale contextual interactions. Unlike classification, which relies on compact global representations, dense prediction tasks such as object detection and semantic segmentation require the preservation of fine spatial details and the modeling of intricate multi-scale interactions. These distinct structural requirements have motivated task-specific adaptations of NAS. In object detection, DetNAS \cite{chen2019detnas} and FNA++ \cite{fang2020fna++} demonstrated that the search for backbones tailored to multi-scale feature fusion produces robust performance. For semantic segmentation, Auto-DeepLab \cite{liu2019auto} exemplifies this trend via the joint optimization of cell-level operators and network-level topologies, which balances spatial precision with global contextual modeling. Furthermore, NAS has been successfully extended to MDE, where PTF-EvoMDE \cite{zhang2025efficient} showed that evolutionary search achieves competitive accuracy-efficiency trade-offs without reliance on costly large-scale pretraining. Despite these advancements, most current methods treat features as generic tensors, which ignores the mathematical importance of the frequency domain for edge preservation.

The prevailing reliance on weight-sharing supernets introduces a critical representation collapse that compromises the ranking reliability of architectural candidates within heterogeneous search spaces. While weight-sharing accelerates the search process, the simultaneous optimization of diverse operators often induces gradient interference, which favors low-frequency semantic stability over high-frequency geometric precision. This imbalance leads to ranking inconsistency, which causes the performance of a subnetwork within the supernet to diverge from its standalone capability. This divergence is particularly problematic in highly irregular search spaces, where capacity variations among subnetworks exacerbate feature modeling imbalances. Consequently, architectures that perform well on 2D benchmarks may fail to capture the high-frequency boundaries necessary for 3D rendering tasks. To address this limitation, a robust alignment strategy is required to synchronize representations across the architectural spectrum.Contemporary developments in domain-specific NAS emphasize the necessity for architectures that capture universal geometric priors rather than task-overfitted features. Specialized applications demonstrate the flexibility of NAS in adapting to unique structural and spectral constraints. Specifically, AdaptorNAS \cite{ang2023adaptornas} utilizes a perturbation-based strategy to derive task-adaptive decoders for arbitrary encoders, which enables efficient model customization. Similarly, gradient-based NAS frameworks for remote sensing incorporate progressive search strategies to enhance robustness and search diversity \cite{peng2020efficient}. In the context of text recognition, specialized search algorithms integrate 3D spatial modeling with transformer-based sequences \cite{zhang2022searching}. For crowded-scene pedestrian detection, NAS-PED \cite{tang2024ped} utilizes CNN-ViT hybrid search spaces to address occlusion and boundary ambiguity. Collectively, these studies illustrate that while NAS is versatile, the primary objective remains the discovery of architectures that generalize across the 2D-to-3D visual spectrum by anchoring representations in both spatial and frequency domains.

\subsection{ENAS}

ENAS provides a robust mathematical framework for navigating the non-differentiable and irregular search spaces inherent in hybrid VSS-ViT architectures. Unlike gradient-based NAS methods that rely on continuous relaxation and impose differentiability constraints, ENAS performs discrete search directly over heterogeneous operators, which avoids the optimization bias introduced by continuous approximations. Consequently, this population-based mechanism maintains architectural diversity across generations, thereby supporting global exploration and mitigating the risk of premature convergence in complex geometric manifolds. Furthermore, ENAS enables a principled formulation for multi-objective optimization through Pareto-based selection. This formulation simultaneously addresses predictive accuracy and hardware-specific constraints to characterize the trade-offs required for deployment in resource-constrained vision scenarios.

However, the structural flexibility of ENAS in these irregular spaces introduces a reliability risk termed ``Representation Collapse,'' where the loss of high-frequency structural information during weight-sharing supernet training leads to a breakdown in subnetwork ranking consistency. While previous advances in multi-objective NAS, such as NAT by Lu et al. \cite{lu2021neural} and SMEMNAS by Xue et al. \cite{xue2026pairwise}, have improved the systematic handling of conflicting objectives and pairwise preferences, they frequently neglect the preservation of underlying geometric features. For 3D-oriented tasks, the fitness estimate of a subnetwork is inherently noisy if the weight-sharing protocol blurs critical boundaries. To address this, current research necessitates cross-task reusability strategies like MTNAS \cite{zhou2023toward}. Nevertheless, these frameworks still struggle without explicit dual-domain alignment to lock high-frequency geometric priors against representation divergence.Extending this rigor to the evaluation phase, achieving an unbiased evolutionary fitness landscape requires moving beyond latency proxies that are system-agnostic toward evaluation engines that are hardware-isolated to eliminate computational noise. Traditional ENAS methods often overlook the latency jitter caused by kernel interference during parallel evaluation. These methods include the variable-length encoding in EMO-SNAS \cite{song2025evolutionary} and the surrogate-based forecasting in CFMOGP-NAS \cite{cao2025comprehensive}. Consequently, the reliance on FLOPs or unshielded measurements introduces distortions into the search trajectory. To ensure that the discovered architectures generalize to 3D rendering that is high-fidelity, it is imperative to utilize a distributed engine that treats hardware isolation as a core component of the search process, thereby guaranteeing that the Pareto-optimal selections reflect physical performance rather than transient system fluctuations.\subsection{Efficient ENAS}
ENAS for vision tasks that are multi-dimensional is bottlenecked by the tension between evaluation throughput and the requirements of geometric representation. Tasks such as 3DGS and MDE require the preservation of fine-grained boundaries, which necessitates a reliable fitness estimate to reflect performance potential. However, obtaining such estimates typically requires substantial training, particularly in vision tasks that are high-resolution and densely supervised \cite{liu2021survey}. Consequently, the overall search process is frequently dominated by evaluation overhead, which substantially limits scalability in expansive architectural search spaces. To address this challenge, current research attempts to accelerate the search process through performance estimation proxies or mechanisms involving weight-sharing.

Although NAS methods that are surrogate-assisted or training-free significantly reduce search overhead, they frequently fail to approximate the performance landscapes of architectures designed for the extraction of high-frequency features. To enhance surrogate reliability, Li et al. \cite{li2022neural} constructed a proxy model of the validation-loss landscape by leveraging intermediate training signals and architectural representations that are graph-based. An alternative direction explores proxies that are training-free. For instance, Zhou et al. \cite{zhou2024training} proposed T-Razor, which evaluates Transformers using synaptic diversity and saliency metrics. Jiang et al. \cite{jiang2025accelerating} introduced MeCo and FLASH, which compute Pearson correlation matrices of feature maps to accelerate operation scoring. Nevertheless, these proxies rely on statistical correlations that often ignore the spectral properties of the feature maps, thereby becoming geometrically blind in vision search spaces that are high-dimensional.

One-shot NAS mitigates training costs via supernets that use weight-sharing, yet it introduces a representation collapse that disrupts the ranking consistency of subnetworks across heterogeneous 2D and 3D domains. While this paradigm significantly reduces computational cost, shared parameters often fail to reflect the representational capacity of individual subnetworks because geometric details that are high-frequency are frequently averaged across different topologies. To mitigate this issue, Chen et al. \cite{chen2023mngnas} proposed MNG-NAS, which enhances supernet training through supervision by multiple teachers. Wang et al. \cite{wang2023dna} introduced DNA, which leverages subnet families guided by distillation to achieve more reliable estimation. Furthermore, Zhong et al. \cite{zhong2025dual} presented DCNAS for image super-resolution, which incorporates a pairwise ranking predictor to align supernet estimates with fully trained performance.To restore ranking reliability within efficient NAS frameworks, it is imperative to move beyond standard weight-sharing to enforce explicit structural and spectral alignment across the supernet. If the supernet is constrained to maintain consistency in both spatial and frequency domains, the fitness landscape becomes smoother and more predictable, which prevents the erosion of sharp boundaries during co-adaptation. This theoretical necessity drives the integration of dual-domain constraints, thereby ensuring that subnetworks maintain geometric precision regardless of their topological variation. Such alignment is critical for capturing universal geometric priors that generalize from 2D pixels to 3D rendering.

Beyond algorithmic efficiency, the integrity of the evolutionary search is further threatened by latency jitter and computational noise inherent in modern parallel hardware evaluation. Most NAS frameworks assume that hardware measurements are stable. However, GPU resource contention in distributed systems creates biased latency data, which misleads the Pareto selection process of the evolutionary algorithm. Consequently, a hardware-isolated evaluation engine is required to eliminate such computational noise and provide an unbiased fitness landscape for the discovery of geometry-aware architectures. This systemic isolation ensures that the discovered architectures are Pareto-optimal in physical environments.

\section{EvoNAS Unified Visual Architecture}
\label{sec:architecture}

The EvoNAS architecture is a hybrid system with a multi-hierarchical structure. This system is designed to reconcile state-space computational efficiency with the global representational capacity required for 3D geometric consistency. The structural topology comprises a supernet encoder using a hybrid VSS-ViT design and a Spatial-Mamba decoder, which collectively facilitate the preservation of high-frequency boundaries. Consequently, the searchable VSS-ViT manifold is formalized through a structured search space defined across multiple architectural dimensions. Such a unified design establishes the rigorous foundation necessary for PST and subsequent multi-objective evolutionary optimization.

\subsection{Architecture Details}\FigureFOne

The architecture of EvoNAS is engineered as a multi-scale framework to bridge the requirements of 2D dense prediction and 3DGS. This framework consists of three principal components: a supernet encoder that utilizes a hybrid VSS and ViT topology, a decoder based on Spatial Mamba, and modular heads for specific tasks. An overview of this organization is illustrated in Fig.~\ref{sec:framework}. The pipeline begins with a Patch Embedding module, which transforms the input image into a high-channel representation with low spatial resolution. This stage functions as a structural geometric quantization step. This step reduces spatial computational overhead and increases channel expressiveness to establish a feature foundation for subsequent stages.

The encoder adopts a phased hybrid topology to mitigate representation collapse during weight-sharing NAS. This topology transitions from local spatial modeling to global structural alignment. The early and intermediate stages comprise Evo-Mamba blocks, which are searchable. These blocks leverage linear-time propagation to capture the geometric continuity of feature maps with high resolution. As the network depth increases, progressive downsampling facilitates semantic abstraction, which leads to a final Evo-Transformer stage. This configuration strengthens cross-region contextual reasoning and aligns the representational structure of the supernet with the Transformer teacher during dual-domain distillation. Consequently, the encoder incorporates a Strip Pooling module \cite{hou2020strip}. This component acts as a directional geometric filter that performs aggregation along horizontal and vertical axes, thereby capturing orthogonal structural priors while preserving deployment efficiency.

The restoration of geometric boundaries is achieved through a decoding mechanism that functions as a global reconstruction engine. Traditional convolutional decoders frequently exhibit limited receptive fields. They also struggle to preserve global structural coherence \cite{yuan2022neural,shao2023iebins,piccinelli2023idisc}. To address these limitations, we construct the decoder using Spatial Mamba blocks \cite{xiao2024spatial}, which we designate as the \emph{Spatial-Mamba Decoder}. In this configuration, the upsampled feature map is channel-compressed and subsequently processed by a Spatial Mamba block. This block models long-range spatial dependencies while simultaneously preserving structural details. These refined features are then integrated with corresponding encoder features via skip connections. This fusion process operates as a geometric refinement step, which effectively reduces boundary ambiguity and enhances the structural integrity of the final reconstruction.The resulting feature maps provided by this encoder-decoder pair serve as a robust geometric foundation. This foundation remains adaptable to a spectrum of downstream tasks through modular task-specific heads. Specifically, the architecture supports optional plug-in modules for object detection \cite{he2017mask}, semantic segmentation \cite{xiao2018unified}, and MDE \cite{godard2019digging}. These task-specific components operate on the multi-scale decoded feature maps and can be seamlessly integrated without modifying the underlying backbone structure. This decoupling allows the evolutionary search to explore a unified optimization space, thereby identifying a single set of geometry-aware weights that remain robust across heterogeneous visual dimensions. Furthermore, this modularity enables the 2D-searched encoder to be hot-swapped into 3D rendering pipelines for novel view synthesis, which confirms its utility as a transferable architectural paradigm for geometric vision.

\subsection{VSS-ViT SuperNet}
\label{sec:vssvit}

To facilitate the extraction of universal geometric priors across the 2D-to-3D spectrum, we conceptualize the encoder as a hybrid representation manifold. This manifold balances linear-time local sensitivity with global semantic consistency. The encoder serves as the foundation for the representational capacity of the vision model. In this role, it captures global structural relationships while it preserves fine-grained local details. Since different vision tasks prioritize these properties to varying extents, an adaptable encoder is required to adjust its depth, capacity, and modeling behavior. To address this requirement, we construct a hybrid VSS-ViT supernet as the searchable backbone of the encoder. The extraction of fine-grained geometry is operationalized through the selective state-space dynamics of the VSS blocks, which are utilized to maintain high-frequency structural boundaries.

The integration of VSS blocks provides a selective memory mechanism that is tuned to high-frequency structural boundaries. This mechanism is critical for geometry-aware tasks. The VSS component is instantiated by utilizing Evo-Mamba blocks, which are derived from selective state-space models that achieve long-range dependency modeling with linear-time computational complexity. Unlike Transformers, which primarily rely on global similarity-based attention mechanisms, Mamba decomposes information propagation into three functional elements. Specifically, an input content term $X$ determines what information is written into the state, a write gate $B$ modulates how the content updates the hidden state, and a read gate $C$ controls how stored information is retrieved for subsequent computation. This write-state-read formulation induces dynamic and selective memory behavior, which enables efficient global context modeling while it maintains localized structural sensitivity.

Complementing this localized structural sensitivity, the terminal layers of the supernet synthesize local features into a coherent global representation. Consequently, a ViT block is strategically incorporated at the final encoder stage to anchor global semantic cohesion and ensure structural compatibility with high-capacity teacher models during dual-domain distillation. This component aligns the representational structure of the supernet with that of the Transformer-based teacher model, which facilitates the CA-DDKD process. By providing a global receptive field at the deepest level, the ViT block ensures that the high-frequency geometric features captured by the preceding VSS blocks are integrated into a semantically robust manifold.The synergy between VSS and ViT components is modulated by a multi-dimensional search space that defines the ultimate capacity of the geometry-aware encoder. We define a four-dimensional search space, which allows the evolutionary algorithm to optimize the granularity of geometric reconstruction under varying hardware constraints. Each dimension is specified as follows:

\begin{itemize}
    \item \textbf{State Dimension ($\text{D}_{\text{STATE}}$):}  
    This parameter controls the dimensionality of the hidden state and directly affects the projection sizes of the write and read gates. A larger state dimension expands the memory capacity for modeling long-range dependencies, but it incurs higher computational overhead. The candidate values are $\{16, 32, 48, 64\}$, which cover configurations from lightweight to high-capacity regimes.

\item \textbf{SSD Expansion Factor ($\text{SSD}_{\text{EXPAND}}$):}  
    This factor determines the expansion applied to the input content term $X$, thereby regulating the richness of information injected into the state. Larger expansion ratios improve the capability of the model to encode fine-grained textures and complex structural boundaries, whereas smaller values reduce redundancy. The search candidates are $\{0.5, 1, 2, 3, 4\}$.

\item \textbf{MLP Expansion Ratio ($\text{MLP}_{\text{RATIO}}$):}  
    The MLP following the state update performs nonlinear transformation and feature recomposition. Its expansion ratio defines the hidden dimensionality of the MLP, which directly influences representational flexibility. Larger ratios enhance the modeling of intricate geometric relationships, while smaller ratios yield more compact architectures. The candidate set is $\{0.5, 1.0, 2.0, 3.0, 3.5, 4.0\}$.

\item \textbf{Stage Depth ($\text{DEPTH}$):}  
    This parameter specifies the number of blocks within each encoder stage. Increasing the depth strengthens hierarchical abstraction and progressive feature refinement. By allowing variable depth configurations, the search framework adapts model complexity to specific deployment constraints, which ensures a favorable balance between computational efficiency and global semantic modeling.
\end{itemize}\section{GPU-Pooled Multi-Objective Evolutionary Search}
\label{sec:NAS}

The optimization of {EvoNAS} is formulated as a geometry-preserving trajectory within a non-convex fitness landscape, which aims to mitigate the representation collapse inherent in weight-sharing search spaces. To stabilize the initial convergence of heterogeneous \textit{VSS-ViT} subnetworks, we introduce \textit{progressive supernet training (PST)}, which decouples the optimization complexity of diverse operator dynamics. Subsequently, the \textit{CA-DDKD} strategy anchors ranking consistency by enforcing structural and spectral alignment, thereby preserving high-frequency geometric priors. To ensure precise multi-objective trade-offs, the \textit{DMMPE} engine utilizes hardware isolation to provide unbiased physical latency measurements, which facilitates the robust extraction of the Pareto-optimal frontier.\subsection{PST for the VSS-ViT Encoder}
\label{sec:progressive_supernet_training}

PST is a fundamental mechanism to mitigate representation collapse and ensure ranking consistency across the heterogeneous VSS-ViT search space. The joint optimization of diverse architectural configurations from scratch often introduces severe gradient interference among shared parameters. This interference results in unstable training dynamics and a degraded performance correlation between the supernet and standalone models \cite{cai2020once}. To address these issues, we implement a curriculum-based activation strategy that warm-starts the search space by optimizing high-capacity configurations before introducing smaller variants. This strategy ensures that the fitness landscape remains smooth for the evolutionary engine, thereby facilitating the discovery of architectures that preserve geometric integrity.

The supernet is initialized by optimizing the maximal configuration to establish a high-fidelity geometric manifold that serves as a reference for all descendant subnetworks. This maximal configuration constitutes the most expressive topology in the search space. The maximal encoder utilizes the largest available state dimension $D_{\text{STATE}} = 64$, the maximum MLP expansion ratio $\text{MLP}_{\text{RATIO}} = 4.0$, and the maximum SSD expansion factor $\text{SSD}_{\text{EXPAND}} = 4$. We establish a robust representational baseline by optimizing this configuration first, which captures the complex 3D rendering and 2D boundary priors necessary for geometry-aware vision. Consequently, this maximal network functions as a geometric teacher that provides a stable parameterization. This parameterization is subsequently inherited by smaller architectures during the progressive expansion process.

We implement a dimension-wise unlocking sequence paired with shape-aligned weight inheritance to preserve structural integrity during capacity reduction. We initiate the reduction of the state dimension by first activating the subset $\{48, 64\}$ once the maximal configuration converges. The $D_{\text{STATE}} = 48$ subnetwork inherits its weights from the $D_{\text{STATE}} = 64$ configuration via a mathematical slicing operation. This operation retains the first $48$ channels of state-related parameters and their corresponding linear projections. This inheritance mechanism ensures that the learned geometric knowledge of the high-capacity model is physically embedded into the low-capacity variants. The full set $\{16, 32, 48, 64\}$ is activated following an adaptation phase of $T_{\text{adapt}}$ epochs to stabilize these new dimensions, thereby allowing the supernet to accommodate a broad spectrum of memory and computational constraints.

Stability in the shared-weight manifold is maintained through a cyclic transition between constrained adaptation for newly activated dimensions and balanced joint optimization for the entire active set. Optimization is restricted to the parameters associated with newly introduced configurations during the adaptation phase $T_{\text{adapt}}$. This restriction minimizes stochastic oscillation in the frozen backbone. A joint optimization phase lasting $T_{\text{joint}}$ epochs is subsequently conducted using balanced uniform sampling across all active configurations. This sampling strategy ensures that each architecture size receives an equivalent update frequency, thereby preventing the supernet from developing a bias toward specific capacities. This balanced optimization is critical for maintaining an unbiased measurement of subnetwork latency, thereby ensuring that the shared weights remain equally compatible with both narrow and wide architectural paths.The expansion process systematically unlocks the expansion ratio of the MLP and the expansion factor of the SSD to refine the capacity for non-linear transformation. We first activate the subset $\{3.0, 3.5, 4.0\}$ for the MLP ratio while the state dimension remains at its expanded state. Newly introduced configurations for the MLP inherit weights of the hidden layers through the truncation of corresponding channels, which ensures structural compatibility and minimizes parameter mismatch. This activation is followed by the inclusion of the complete set $\{0.5, 1.0, 2.0, 3.0, 3.5, 4.0\}$. Similarly, the SSD expansion factor is increased by activating the subset $\{2, 3, 4\}$, which utilizes aligned slicing along the expanded pathway for content. By performing targeted updates on these newly activated pathways, we facilitate the alignment of feature statistics across varying expansion factors, thereby stabilizing the optimization for the flow of geometric features.

The final stage of PST unlocks variable depth for each stage through block inheritance based on prefixes, which ensures that feature propagation remains consistent as the computational path length fluctuates. Depth inheritance is achieved by reusing the leading blocks of the configuration with maximal depth for all shallower subnetworks, while skipped blocks are implemented as identity mappings. This reuse based on prefixes is essential for maintaining the alignment of features with high frequency across different architectural depths, thereby preventing the loss of sharp geometric boundaries during the search. After enabling variable depth, a final convergence phase of $T_{\text{final}}$ epochs is performed with comprehensive joint optimization across all dimensions. This final alignment ensures that the supernet provides a reliable proxy for the standalone performance of any sampled subnetwork, thereby establishing a consistent fitness landscape for the subsequent architecture search that is geometry-aware.

\subsection{Two-Stage VSS-ViT Supernet Optimization via CA-DDKD}

The optimization of the VSS-ViT supernet follows a bifurcated trajectory that decouples the stability of general-purpose features from geometric refinement for specific tasks. In the primary stage, the supernet undergoes extensive pre-training on ImageNet via the PST strategy. By initializing from the maximal configuration and incrementally expanding the searchable structural dimensions, the supernet establishes a foundation of stable and transferable shared weights. This hierarchical process ensures representational compatibility across heterogeneous architectural paths, which provides a robust initialization for downstream adaptation. Consequently, the second stage adapts the pre-trained supernet to datasets from the target domain through the CA-DDKD framework. Under a consistent protocol for progressive sampling, the system jointly optimizes sampled subnetworks using task-specific supervision and alignment guided by a teacher, thereby encoding inductive priors relevant to the task while preserving consistency across architectures.Standard weight-sharing NAS for dense prediction tasks is limited by the representation collapse of high-frequency geometric features, which manifests as ranking inconsistency during subnetwork evaluation. Diverse downstream vision tasks impose heterogeneous requirements, such as global contextual reasoning and multi-scale semantic aggregation. Although VSS-based subnetworks offer substantial computational efficiency due to their linear-time complexity, relying exclusively on sparse task-level supervision during training often results in suboptimal shared representations. Specifically, limited task annotations provide insufficient constraints on long-range geometric dependencies. Moreover, capacity variations induced by subnetwork sampling introduce an imbalance in high-frequency feature modeling. This imbalance adversely affects boundary precision and texture reconstruction. Consequently, these factors degrade the quality of shared parameters and distort the fitness landscape, which necessitates a stabilizing mechanism to preserve ranking reliability.

We introduce CA-DDKD to anchor the supernet weight-sharing manifold to a geometry-aware teacher space, which ensures structural alignment across diverse architectural configurations. We adopt Depth Anything \cite{yang2024depth} as the teacher model due to its Transformer-based architecture and large-scale pre-training. These properties together encode robust geometry-aware priors. During optimization, every subnetwork sampled under the progressive protocol is supervised by both ground-truth labels and dual-domain teacher guidance. In the spatial domain, intermediate feature representations are aligned to encourage consistent structural layouts. Simultaneously, spectral alignment is enforced in the frequency domain to preserve global structures while enhancing high-frequency boundary details. This dual-domain distillation framework regularizes representation learning at the feature level, which mitigates distribution discrepancies among heterogeneous subnetworks and improves the transferability of the learned encoder.

\FigureFTwo

To implement this refinement, the PST schedule is maintained during the second-stage adaptation where each subnetwork functions as a student model. Knowledge distillation is integrated into the shared-weight optimization process rather than functioning as a secondary auxiliary. When a new architectural dimension is activated, the corresponding subnetworks inherit parameters from structurally larger configurations through a progressive inheritance strategy. Immediately following this inheritance, the subnetworks are constrained by teacher outputs in both spatial and frequency domains, which accelerates the stabilization of predictive distributions. This teacher-guided mechanism mitigates optimization instability during dimension expansion and reduces representational divergence under shared weights. During the final convergence phase, all subnetworks are jointly trained with balanced sampling to ensure that shared parameters remain coherent across architectural variations, thereby improving subnetwork comparability for evolutionary search.Structural compatibility is further enforced by the architectural pipeline illustrated in Fig.~\ref{sec:supernet training}, where hierarchical features produced by VSS stages are first transformed by a terminal ViT block. This transformation yields representations that are compatible with the ViT architecture. Specifically, the token dimensionality and structural organization of these representations are aligned with the teacher model. The resulting student tokens are subsequently processed by a projection module based on multi-head attention, which utilizes learnable query embeddings to match the spatial resolution of the teacher. In the spatial domain, semantic alignment on a global scale is enforced by minimizing the mean squared error (MSE) between corresponding tokens. To preserve geometric boundaries, the framework projects features into the spectral domain via the DCT, thereby isolating and supervising high-frequency components that are typically suppressed during weight sharing. By suppressing the dominant direct-current (DC) component and applying an additional MSE constraint on representations enhanced in the frequency domain, the student subnetworks learn to simultaneously capture geometric structure at a global scale and refined details at the local level.

The final optimization objective integrates supervision via ground truth, pseudo-labels induced by the teacher, and constraints from dual-domain distillation into a unified formulation. This formulation guarantees a smooth fitness landscape for the training process. At the decoder stage, student tokens are propagated through task-specific heads for MDE or semantic segmentation, which are supervised by task-dependent losses. In parallel, the teacher model generates refined predictions that function as pseudo-labels, and a consistency loss is imposed between teacher and student predictions. This soft supervision transfers the global contextual reasoning of the teacher to the student, particularly in regions where annotations are sparse or structurally complex. By combining distillation across dual domains at the feature level with dual supervision at the prediction level, the framework establishes a coherent optimization strategy. This approach reduces performance variance across heterogeneous subnetworks and enhances the generalization capability of the supernet with shared weights.

The unified landscape for multi-objective optimization is mathematically formalized as follows:\begin{equation}
\mathcal{L} = (1-\theta)\mathcal{L}_{\mathrm{gt}} 
+ \theta \mathcal{L}_{\mathrm{pseudo}} 
+ \alpha_{1}\mathcal{L}_{\mathrm{spat}} 
+ \alpha_{2}\mathcal{L}_{\mathrm{freq}},
\label{eq:total_loss}
\end{equation}The total training objective integrates several loss components to ensure robust supernet optimization. In this formulation, $\mathcal{L}_{\mathrm{gt}}$ denotes the task-specific loss derived from ground-truth annotations, such as cross-entropy for classification or segmentation and $\ell_1$ regression for depth estimation and scale-invariant logarithmic (SILog) loss for MDE. The term $\mathcal{L}_{\mathrm{pseudo}}$ represents the prediction-level consistency loss induced by teacher outputs, while the balancing coefficient $\theta \in [0,1]$ regulates the trade-off between hard supervision and soft guidance. Furthermore, the terms $\mathcal{L}_{\mathrm{spat}}$ and $\mathcal{L}_{\mathrm{freq}}$ correspond to spatial-domain and frequency-domain feature alignment losses, respectively. The hyperparameters $\alpha_1$ and $\alpha_2$ regulate the relative influence of structural alignment and spectral refinement within the distillation framework.

The distillation component leverages these complementary constraints to maximize ranking stability. Specifically, $\mathcal{L}_{\mathrm{spat}}$ enforces global semantic organization, while $\mathcal{L}_{\mathrm{freq}}$ emphasizes edge structures that are often underrepresented in purely spatial supervision. The relative strengths of these alignments are controlled by $\alpha_{1}$ and $\alpha_{2}$, which enables flexible refinement of high-frequency geometric details. By embedding these objectives into the progressive refinement process, the ViT-based teacher provides stable supervision across all subnetwork capacities. This mechanism effectively mitigates representation misalignment under shared weights, thereby preserving structural compatibility. Consequently, distillation-enhanced optimization improves subnetwork comparability and strengthens ranking consistency. This improvement ensures that validation-time supernet evaluation is highly predictive of standalone performance, which enables reliable fitness estimation during the subsequent evolutionary search.\subsection{Distributed Multi-Model Parallel Evaluation Framework}
\label{sec:distributed_eval}

The reliability of multi-objective evolutionary search is fundamentally contingent upon the precision of the fitness landscape, which is frequently compromised by stochastic computational noise and latency jitter during concurrent hardware-aware evaluation. In dense prediction tasks, evaluating high-resolution candidates imposes substantial GPU loads, which frequently results in kernel interference when multiple models compete for Streaming Multiprocessors (SMs). This resource contention results in distorted latency measurements, which misguide the evolutionary algorithm, thereby potentially discarding superior architectures due to transient hardware bottlenecks rather than intrinsic structural inefficiencies. Consequently, the search process requires a robust mechanism to decouple architectural performance from systemic execution variability to maintain the integrity of the Pareto-optimal frontier. To address this challenge, the evaluation phase must be treated as a controlled experimental environment that ensures measurement repeatability across diverse hardware configurations.To mitigate this measurement bias, we introduce the DMMPE framework, which serves as a hierarchical orchestration engine to enforce hardware isolation and temporal determinism across heterogeneous GPU clusters. This framework is structured into a three-tier hierarchy: cross-GPU task scheduling, intra-GPU multi-process coordination, and intra-process multi-model execution. Rather than functioning solely as an acceleration tool, this system acts as a high-fidelity sensor for the evolutionary agent, which ensures that every fitness score reflects the intrinsic geometric capacity of the architecture instead of systemic overhead. By separating architectural instantiation from performance profiling, the framework reduces sequential dependencies and alleviates the impact of resource fragmentation. This hierarchical organization transforms candidate evaluation from a stochastic bottleneck into a component of the evolutionary pipeline that is both scalable and high-throughput, as illustrated in Fig.~\ref{sec:EvoX}.

At the macroscopic level, DMMPE employs a decoupled Master-Worker architecture to eliminate global synchronization barriers and ensure load-balanced architectural dispatch. The master process first decodes the population $\mathcal{P}$ into executable architectural configurations $\{c_i\}_{i=1}^{N}$ and partitions them into model pools $\{\mathcal{M}_k\}$, each containing $B_m$ candidates. These model pools are dispatched to worker processes distributed across the GPU set $\mathcal{G}$ using a producer-consumer paradigm, which allows workers to retrieve tasks and perform validation independently. Hardware metrics, including inference latency and multiply-accumulate operations (MACs), are measured in isolation on the assigned GPU to prevent cross-device interference. This asynchronous strategy eliminates the "straggler effect," where slower models delay the entire generation, thereby providing stable scalability across heterogeneous hardware environments. Predictive performance is simultaneously evaluated using the trained supernet $S$ on the validation dataset $\mathcal{D}_{\mathrm{val}}$, which results in a multi-objective fitness matrix that guides the subsequent selection process.

To maintain high throughput without sacrificing measurement fidelity, the framework utilizes NVIDIA Multi-Process Service (MPS) to facilitate deterministic kernel interleaving within shared CUDA contexts. Within each GPU, multiple worker processes operate under a shared primary context, which significantly reduces the overhead associated with frequent context switching and memory management. Since subnetwork evaluation often involves lightweight and fragmented kernel launches, this shared-context model effectively consolidates scattered computational workloads into a continuous execution stream. Furthermore, at the process level, worker nodes utilize a unified data loading and preprocessing pipeline that amortizes fixed overheads across multiple candidate models. This consolidation stabilizes the latency signal, which ensures that the measured execution time is strictly proportional to the structural complexity of the model rather than Python-level dispatch delays or I/O bottlenecks. Consequently, GPU occupancy is maximized even when individual subnetworks are computationally sparse, thereby preserving high validation throughput while maintaining the precision required for fine-grained architecture comparison.DMMPE transforms the evaluation phase from a stochastic bottleneck into a high-fidelity and deterministic pipeline, which enables the evolutionary process to converge on architectures that represent physical Pareto optimality. Specifically, the precision afforded by hardware isolation allows the search algorithm to distinguish between models that appear efficient due to transient resource availability and those that possess geometric efficiency. This measurement integrity is a prerequisite for the effective operation of the CA-DDKD strategy, as it ensures that the frequency-domain constraints are applied to architectures with accurately characterized latency profiles. Consequently, the DMMPE framework provides the empirical data necessary for the discovery of universal geometric priors, which facilitates the transition of the searched encoder from 2D dense prediction to 3D rendering tasks. Through this systematic orchestration, the framework establishes a reliable foundation for exploring the VSS-ViT search space without the confounding effects of hardware-induced noise.\begin{algorithm}[t]
\caption{Distributed Multi-Model Parallel Evaluation}
\label{alg:parallel-eval}
\begin{algorithmic}[1]
\item[] \textbf{Input:} Population $\mathcal{P}=\{g_i\}_{i=1}^{N}$; trained supernet $\mathcal{S}$; validation set $\mathcal{D}_{val}$; processes per GPU $N_p$; models per process $B_m$; GPUs $\mathcal{G}$.
\item[] \textbf{Output:} Fitness matrix $F \in \mathbb{R}^{N \times 3}$: validation error $\varepsilon$, latency $\tau$, MACs $m$.
\STATE $\{\mathbf{c}_i\}_{i=1}^{N} \gets \text{Decode}(\mathcal{P})$
\STATE Partition $\{\mathbf{c}_i\}$ into model pools $\{\mathcal{M}_k\}$ of size $B_m$
\FOR{$g, p \in \mathcal{G} \times \{1, \dots, N_p\}$}
    \FOR{each model pool $\mathcal{M}_k$ assigned to process $(g,p)$}
        \STATE $\varepsilon[\mathcal{M}_k] \gets \text{EvalErr}(\mathcal{S}, \mathcal{D}_{val}, \mathcal{M}_k)$
    \ENDFOR
\ENDFOR
\FOR{$g \in \mathcal{G}$}
    \FOR{each config $\mathbf{c}_i$ assigned to GPU $g$}
        \STATE $(\tau_i, m_i) \gets \text{Measure}(\mathcal{S}, \mathbf{c}_i)$
        \STATE $\tau[i] \gets \tau_i$
        \STATE $m[i] \gets m_i$
    \ENDFOR
\ENDFOR
\STATE $\mathcal{J}(\mathcal{P})[:,1] \gets \varepsilon$; $\mathcal{J}(\mathcal{P})[:,2] \gets \tau$; $\mathcal{J}(\mathcal{P})[:,3] \gets m$
\RETURN $\mathcal{J}(\mathcal{P})$
\end{algorithmic}
\end{algorithm}\subsection{EvoNAS pipeline}
\label{sec:evoxnas_pipeline}

\FigureFThree
\renewcommand{\algorithmiccomment}[1]{\hfill $\triangleright$ #1}\begin{algorithm}[t]
\caption{EvoNAS pipeline}
\label{alg:prog-train-search}
\begin{algorithmic}[1]
\item[] \textbf{Input:} ImageNet dataset $\mathcal{D}_{IN}$, target dataset $\mathcal{D}_{task}$; validation set $\mathcal{D}_{val}$;
progressive schedule $\Phi=\{\phi_p\}_{p=0}^{P}$ with candidate sets $\{\mathcal{C}_p\}$ and iterations $\{T_p\}$, where $\mathcal{C}_0=\{g_{\max}\}$ and $\mathcal{C}_p\subseteq\mathcal{C}_{p+1}$;
teacher network $\mathcal{T}$;
learning rate $\eta$;
population size $N_{\text{pop}}$ and generations $G$.
\item[] \textbf{Output:} Pareto-optimal set $\mathcal{A}^\star$ of (genotype, objectives).
\STATE Initialize supernet $\mathcal{S}$ with parameters $\theta$ and heads $\{h_{\text{cls}},h_{\text{task}}\}$.
\FOR{each stage $\textsc{Stage} \in \{\textsc{Pretrain}, \textsc{Finetune}\}$}
    \IF{$\textsc{Stage}$ is \textsc{Pretrain}}
        \STATE $\mathcal{D} \gets \mathcal{D}_{IN}$; $h \gets h_{\text{cls}}$
    \ELSE
        \STATE $\mathcal{D} \gets \mathcal{D}_{task}$; $h \gets h_{\text{task}}$
    \ENDIF
    \FOR{$p \gets 0$ \textbf{to} $P$}
        \FOR{$t \gets 1$ \textbf{to} $T_p$}
            \STATE Sample minibatch $x \sim \mathcal{D}$
            \STATE Sample genotype $g \sim \mathcal{C}_p$
            \STATE $(\hat{y}_S, f_S) \gets \mathcal{S}(x;g,h)$
            \IF{$\textsc{Stage}$ is \textsc{Pretrain}}
                \STATE $\ell \gets \mathcal{L}_{\text{cls}}(\hat{y}_S)$
            \ELSE
                \STATE $\ell \gets \mathcal{L}_{\text{task}}(\hat{y}_S) + \mathcal{L}_{\text{pseudo}}(\hat{y}_S) + \mathcal{L}_{\text{spat}}(f_S, \mathcal{T}(x)) + \mathcal{L}_{\text{freq}}(f_S, \mathcal{T}(x))$
            \ENDIF
            \STATE $\theta \gets \theta - \eta \nabla_\theta \ell$
        \ENDFOR
    \ENDFOR
    \ENDFOR
\STATE Initialize population $\mathcal{P}_0$ with $N_{\text{pop}}$ genotypes
\STATE $\mathcal{J}(\mathcal{P}_0) \gets \text{ParallelEval}(\mathcal{P}_0, \mathcal{S}, \mathcal{D}_{val})$\COMMENT{Alg.~\ref{alg:parallel-eval}}
\FOR{$t \gets 0$ \textbf{to} $G-1$}
    \STATE $\mathcal{Q}_t \gets \text{OffspringGeneration}(\mathcal{P}_t)$
    \STATE $\mathcal{J}(\mathcal{Q}_t) \gets \text{ParallelEval}(\mathcal{Q}_t, \mathcal{S}, \mathcal{D}_{val})$\COMMENT{Alg.~\ref{alg:parallel-eval}}
    \STATE $(\mathcal{P}_{t+1}, \mathcal{J}(\mathcal{P}_{t+1})) \gets \text{Survival}(\mathcal{P}_t \cup \mathcal{Q}_t, \mathcal{J}(\mathcal{P}_t) \cup \mathcal{J}(\mathcal{Q}_t))$
\ENDFOR
\STATE $\mathcal{A}^\star \gets \text{ParetoFront}(\mathcal{P}_G, \mathcal{J}(\mathcal{P}_G))$
\RETURN $\mathcal{A}^\star$
\end{algorithmic}
\end{algorithm}We formalize the discovery of geometry-aware backbones as a multi-objective optimization problem that seeks to reconcile predictive fidelity with hardware constraints. Specifically, the proposed EvoNAS pipeline addresses the trade-offs encountered in dense vision tasks for edge-oriented applications by jointly optimizing three objectives: prediction error, inference latency, and computational complexity measured in multiply--accumulate operations (MACs). These objectives are conflicting, because increasing architectural capacity generally improves predictive accuracy but simultaneously elevates computational costs and inference latency. Such increases are often prohibitive in resource-constrained deployment scenarios. To navigate this trade-off, we denote an architecture configuration as $c \in \mathcal{S}$, where $\mathcal{S}$ represents the search space. The optimization objective is defined as follows:\begin{equation}
\min_{c \in \mathcal{S}} 
\quad 
\mathbf{F}(c) 
=
\left[
f_{\mathrm{err}}(c), 
f_{\mathrm{lat}}(c), 
f_{\mathrm{mac}}(c)
\right],
\end{equation}The multi-objective formulation incorporates validation prediction error $f_{\mathrm{err}}(c)$, inference latency $f_{\mathrm{lat}}(c)$ measured on target hardware, and computational complexity $f_{\mathrm{mac}}(c)$. To approximate the Pareto-optimal set, we utilize the NSGA-II algorithm \cite{kalyanmoy2002fast}, which performs non-dominated sorting and crowding-distance-based selection to maintain architectural diversity along the Pareto frontier. This evolutionary mechanism ensures that the resulting population covers a well-distributed spectrum of optimal trade-offs between accuracy and efficiency.

The VSS-ViT supernet follows a phased optimization strategy to establish a stable and transferable feature space by transitioning from general semantic initialization to geometry-specific alignment. As illustrated in Algorithm~2 and Fig.~\ref{sec:EvoX}, EvoNAS optimizes the supernet $S$ with parameters $\theta$ and task-specific heads through ImageNet pretraining followed by target-dataset refinement. During the first stage, the supernet is optimized on the ImageNet dataset to establish a foundational manifold for the hybrid VSS-ViT blocks. At each iteration, a genotype is sampled according to a progressive schedule, and the shared parameters are updated via stochastic gradient descent. This stage produces an initialization that is both robust and transferable, which maintains compatibility across heterogeneous architectural paths.

The CA-DDKD framework provides the theoretical anchor to mitigate representation collapse during the second stage of supernet optimization. During target-dataset refinement, we integrate guidance from a frozen teacher model through the CA-DDKD framework. For each sampled genotype, the refinement objective integrates task-level supervision with dual-domain feature alignment. By enforcing structural alignment in both spatial and frequency domains via the DCT, CA-DDKD explicitly preserves essential high-frequency geometric boundaries. This strategy effectively reduces representation divergence under shared weights, which ensures that subnetwork fitness estimates are reliable proxies for standalone performance. Consequently, the distillation process stabilizes the ability of the supernet to model high-frequency geometry, thereby resolving the ranking inconsistency issue typically found in weight-sharing NAS (Lines~16--18 of Algorithm~2).

EvoNAS employs a hybrid genotype representation to navigate the distinct structural logic of the VSS-ViT search space once the supernet is anchored by dual-domain priors. Each candidate architecture is encoded using integer variables for stage-level hyperparameters and a binary vector for depth configuration. The integer segment parameterizes searchable dimensions, including state dimensions, SSD expansion factors, and MLP ratios, while the binary vector indicates block activation. To maintain structural integrity during evolution, we utilize operator-representation alignment. Specifically, two-point crossover and polynomial mutation are applied to the integer segment to enable bounded local perturbations. Conversely, uniform crossover and bit-flip mutation are applied to the binary depth segment. This differentiated approach preserves structural continuity in stage-level configurations while encouraging diversity in depth patterns, which allows the evolutionary algorithm to effectively traverse the hybrid architectural manifold.The DMMPE engine enforces strict hardware isolation during latency measurement to eliminate stochastic noise from the evolutionary fitness landscape. Standard parallel evaluation on shared GPUs often suffers from kernel interference and resource contention, thereby leading to biased latency measurements that misguide the search. The DMMPE engine, i.e., \textsc{ParallelEval} in Lines~23 and 26 of Algorithm~2, addresses this challenge by utilizing hardware isolation to guarantee unbiased physical measurements. By removing computational noise from the fitness signals, the engine ensures that the evolutionary process is guided by the physical performance of each subnetwork. Consequently, this hardware-isolated evaluation is essential for identifying architectures that are Pareto-optimal in real-world deployment contexts.

The final evolutionary phase utilizes the NSGA-II framework to identify a diverse set of architectures that define the Pareto frontier for geometric tasks. Algorithm~2 initializes a population $\mathcal{P}_0$ with $N_{\mathrm{pop}}$ genotypes (Line~22). At each generation $t$, an offspring population $\mathcal{Q}_t$ is generated via genetic operators, and their objectives $\mathcal{J}(\mathcal{Q}_t)$ are computed using the DMMPE pipeline. Survival selection is subsequently performed on the union $\mathcal{P}_t \cup \mathcal{Q}_t$ through non-dominated sorting to determine the next generation $\mathcal{P}_{t+1}$. After $G$ generations, the final Pareto-optimal set $\mathcal{A}^{\star}$ is extracted. This process yields the ``EvoNet'' family, which consists of architectures that demonstrate notable transferability and capture fundamental physical geometry across both 2D and 3D vision domains.

\section{Experimental settings}
\label{sec:experiments}

This section delineates the experimental framework designed to validate the geometry-aware NAS paradigm across a multi-stage evaluation hierarchy. To establish the transferability of universal visual priors, we curate a benchmark suite that spans four levels of geometric complexity, which ranges from 2D instance localization to high-fidelity 3D rendering. Consequently, the performance is assessed through a tripartite metric space that balances predictive accuracy with unbiased physical efficiency. The training protocol integrates CA-DDKD to enforce spectral alignment during supernet pre-training, which stabilizes the ranking consistency required for evolutionary search. Furthermore, the DMMPE engine utilizes hardware isolation to guarantee reliable latency measurements, thereby eliminating computational noise during the optimization phase.

\subsection{Benchmark Datasets}

We conduct experiments on four representative benchmarks that cover various dense vision tasks and scene domains. These datasets include object detection, semantic segmentation, and MDE across both outdoor and indoor environments. Specifically, COCO evaluates instance-level detection performance in complex scenes characterized by rich object interactions. ADE20K focuses on pixel-wise semantic understanding across diverse scene categories. KITTI represents outdoor depth estimation in autonomous driving scenarios, which exhibit perspective variation at a large scale and depth discontinuities across long ranges. In contrast, NYU Depth v2 evaluates indoor depth estimation under cluttered layouts, short-range structures, and geometric details that are fine-grained. This diversity of tasks, supervision granularity, and environmental conditions enables a comprehensive assessment of the robustness, adaptability, and deployment efficiency of the proposed framework.\paragraph{COCO.}
The COCO dataset~\cite{lin2014microsoft} is a large-scale benchmark for object detection, comprising 80 object categories with complex scenes and instance-level annotations. We adopt the standard COCO 2017 split, using \textit{train2017} for training and \textit{val2017} for evaluation. Images are resized such that the shorter side is 800 pixels while the longer side does not exceed 1333 pixels. This single-scale configuration is applied consistently during both training and inference.

\paragraph{ADE20K.}
ADE20K~\cite{zhou2017scene} is a widely used scene parsing benchmark containing 150 semantic categories. We follow the official train/validation split and report mean Intersection over Union (mIoU) on the validation set. During training, images are randomly resized with a scale factor sampled from $[0.5, 2.0]$, followed by random cropping to $512 \times 512$ and horizontal flipping. For evaluation, images are resized such that the longer side is 2048 pixels and the shorter side is 512 pixels. Single-scale full-image inference is conducted without cropping.

\paragraph{KITTI.}
The KITTI dataset~\cite{geiger2013vision} is an outdoor benchmark for MDE collected in autonomous driving environments. The original image resolution is $375 \times 1242$ pixels. We adopt the widely used Eigen split~\cite{eigen2014depth}, which contains 23,488 training images and 697 test images from the left camera view. During evaluation, we apply the central cropping protocol proposed in~\cite{garg2016unsupervised} to ensure fair comparison with prior work.

\paragraph{NYU-Depth-v2 (NYU v2).}
NYU v2~\cite{silberman2012indoor} is an indoor RGB-D benchmark consisting of approximately 120K image pairs from 464 scenes with a resolution of $640 \times 480$. Following the standard split introduced in~\cite{eigen2014depth}, 249 scenes are used for training and 654 images for testing. Due to synchronization noise between RGB and depth sensors, we retain 36,253 valid training samples. For evaluation, predicted depth maps are center-cropped to $561 \times 427$, consistent with~\cite{eigen2014depth}.

\subsection{Evaluation Metrics and Geometric Fidelity}

To rigorously validate the efficacy of the proposed CA-DDKD in capturing universal geometric priors, we employ a multi-tiered evaluation suite spanning 2.5D depth estimation, 2D dense prediction, and 3D neural rendering. These tasks serve as a rigorous assessment of representation consistency across varying geometric manifold complexities. Consequently, this multi-faceted evaluation strategy ensures that the searched architectures avoid dataset-specific over-fitting.

In tasks sensitive to boundary precision, such as MDE and Semantic Segmentation, we prioritize metrics that penalize high-frequency structural deviations. Specifically, for NYU Depth v2 and KITTI, we adopt Absolute Relative Error (Abs Rel) as the primary metric, which is supplemented by Root Mean Squared Error (RMSE) and the threshold accuracy $\delta_1$~\cite{eigen2014depth}. While Abs Rel provides a normalized estimate of proportional error, RMSE emphasizes large deviations near structural discontinuities, thereby validating the success of spectral alignment in preserving depth edges. For the ADE20K benchmark, we utilize mean Intersection over Union (mIoU)~\cite{everingham2010pascal} to quantify spatial consistency and the preservation of sharp semantic boundaries under complex occlusions.

For object-level reasoning on the MS COCO benchmark, we utilize the AP@[0.5:0.95] metric to jointly assess classification confidence and the geometric fidelity of bounding-box regression~\cite{lin2014microsoft}. This metric integrates performance across varying localization strictness levels, which demonstrates the capacity of the hybrid VSS-ViT search space to fuse global contextual reasoning with local spatial precision. By evaluating the mean Average Precision across ten IoU thresholds, we ensure that the architectural search process remains sensitive to fine-grained localization accuracy rather than coarse object detection alone.

While 2D and 2.5D metrics evaluate surface-level precision, they remain insufficient to prove volumetric consistency. Thus, we extend our evaluation to 3D rendering metrics via novel view synthesis on the RealEstate10K dataset. We report Peak Signal-to-Noise Ratio (PSNR), Structural Similarity Index (SSIM), and Learned Perceptual Image Patch Similarity (LPIPS) to measure rendering fidelity and perceptual quality within the 3DGS framework. These metrics explicitly confirm whether the dual-domain constraints enforced during the 2D search phase effectively translate to the reconstruction of physical 3D volumes with high-frequency texture retention.

Beyond theoretical complexity, we measure physical latency through our hardware-isolated DMMPE engine to ensure a noise-free fitness landscape for the evolutionary search. While model parameters (Params) and multiply–accumulate operations (MACs) provide architecture-level complexity estimates, they remain insufficient for characterizing real-world throughput due to operator-level kernel overheads. To eliminate latency jitter, which is caused by GPU resource contention, our DMMPE engine enforces strict hardware isolation and explicit synchronization before and after timing. Inference latency is measured with FP32 precision under batch size 1 after an extensive warm-up phase, which ensures that the resulting Pareto-optimal architectures are optimized for actual physical deployment. Furthermore, we report Normalized Information Density (NID)~\cite{mousavi2023dass} to quantify the ratio between task performance and parameter count, thereby highlighting the parameter efficiency of our geometry-aware paradigm.

\subsection{Implementation Details}The framework utilizes a DMMPE engine to ensure an unbiased fitness landscape during the evolutionary search. This engine is characterized by strict hardware isolation. The physical infrastructure consists of a cluster equipped with eight NVIDIA RTX 3090 GPUs and dual Intel Xeon Gold 5318Y CPUs. By employing hardware isolation and GPU resource pooling, the engine eliminates the latency jitter typically caused by kernel interference, which provides a controlled environment for architectural benchmarking. Implementation is realized via PyTorch with automatic mixed precision (AMP). This approach maximizes computational throughput and memory efficiency during high-dimensional search.

The stabilization of the hybrid VSS-ViT search space is achieved through a multi-stage protocol for progressive unlocking that mitigates gradient interference within the weight-sharing supernet. Specifically, initial pre-training on ImageNet-1K establishes a transferable foundation using the AdamW optimizer with an initial learning rate of 0.001 and weight decay of 0.05. We employ a linear warm-up for 20 epochs followed by cosine annealing. Simultaneously, we apply label smoothing, DropPath, and diverse data augmentations including RandAugment and CutMix. The optimization begins with the maximal configuration for 300 epochs. Subsequently, the searchable dimensions, specifically \texttt{D\_STATE}, \texttt{MLP\_RATIO}, and \texttt{SSD\_EXPAND}, are sequentially unlocked. Each expansion phase includes a 25-epoch transition period to stabilize newly activated subnetworks, which is followed by a 120-epoch consolidation period to ensure representational compatibility across the expanded manifold.

For downstream adaptation, we anchor the feature representations of the supernet through the CA-DDKD strategy, which enforces spatial and frequency-domain alignment against high-fidelity teacher models. This adaptation is applied to MDE on NYU v2 and KITTI, Semantic Segmentation on ADE20K, and Object Detection on COCO. In MDE, the supernet undergoes sequential unlocking with schedules of $5+20$ epochs per dimension. Specifically, we utilize Depth Anything as the teacher model to provide pseudo-labels and DCT based frequency alignment. Semantic segmentation is implemented within the UPerNet framework~\cite{xiao2018unified}, which employs a crop size of $512 \times 512$ and synchronized batch normalization to stabilize multi-GPU gradients. For object detection, we utilize the Mask R-CNN framework~\cite{he2017mask} with a progressive $2+8$ epoch schedule per dimension, which ensures that the distillation of multi-scale features preserves critical geometric boundaries.

The architectural discovery process employs a constrained NSGA-II algorithm to navigate the non-convex trade-offs between geometric fidelity and physical inference latency. We initialize a population of 96 individuals and execute the evolutionary search for 200 generations with crossover and mutation probabilities of 0.95 and 0.1, respectively. The multi-objective fitness function simultaneously evaluates task-specific accuracy metrics, computational complexity (MACs), and physical latency provided by the DMMPE engine. Following the identification of the Pareto-optimal frontier, standalone architectures are retrained under fixed topologies. These models inherit weights from the converged supernet and undergo task-specific refinement, such as 40 epochs for MDE and 80k iterations for ADE20K. This refinement allows the system to fully realize the capacity of the discovered geometry-aware configurations.To validate the universal geometric priors captured by the searched encoder, the discovered EvoNet is integrated into a 3DGS pipeline for high-fidelity novel view synthesis. This cross-dimensional transfer utilizes the RealEstate10K benchmark to assess the ability of the 2D-searched backbone to provide geometric context for 3D rendering primitives. The encoder replaces conventional backbones and provides features that regularize the initialization of Gaussian components. By maintaining high-frequency spectral alignment during the search phase, the architecture enables the 3DGS framework to achieve high rendering fidelity while maintaining a substantial reduction in parameter overhead. This result confirms the robustness and transferability of the dual-domain representation alignment paradigm.

\section{Experimental Study}
\label{sec:experiments_ss}

{The experimental study is organized to answer four questions: (1) Can EvoNet achieve superior accuracy-efficiency Pareto trade-offs on representative 2D dense prediction tasks? (2) Can monocular depth estimation verify that the proposed CA-DDKD strategy preserves geometry-sensitive representations under weight sharing? (3) Do the architectures discovered in 2D transfer effectively to 3D novel view synthesis? (4) Does the DMMPE engine provide a reliable evaluation protocol for hardware-aware multi-objective search? To answer these questions, the searched architectures (i.e., EvoNet) are evaluated on COCO for object detection, ADE20K for semantic segmentation, KITTI and NYU v2 for monocular depth estimation, and RealEstate10K for 3D novel view synthesis.}

Performance on detection, segmentation, and depth estimation serves as the primary test of whether EvoNet can maintain both predictive quality and geometric fidelity across heterogeneous visual tasks. In particular, monocular depth estimation functions as a geometric stress test, since it directly probes whether the learned features preserve boundary precision, local structure, and pixel-level consistency when the encoder is trained in a weight-sharing supernet. From this perspective, the MDE results are intended not merely to report task performance, but to examine whether CA-DDKD successfully stabilizes geometry-aware representations across subnetworks with different topologies.

To further examine transferability beyond 2D dense prediction, we evaluate the searched EvoNet encoder within a 3DGS pipeline for novel view synthesis on RealEstate10K. This experiment is designed to test whether the geometry-aware search process captures transferable physical structure rather than task-specific 2D patterns. Strong rendering performance in this setting indicates that the searched encoder retains the geometric priors required for 3D reconstruction, while the substantial reduction in model size demonstrates that such transferability can be achieved without relying on large backbone models.

Finally, we investigate whether the search trajectory itself is trustworthy. Since hardware-aware multi-objective optimization is sensitive to latency jitter and measurement noise, the DMMPE engine is evaluated as a reliability mechanism rather than merely a systems component. By enforcing hardware isolation during candidate evaluation on eight NVIDIA RTX 3090 GPUs, DMMPE is intended to reduce runtime interference and provide more stable fitness estimates, so that the resulting Pareto frontier reflects architectural quality rather than incidental execution artifacts.

\Figurenin\subsection{Results on COCO}

\TabCOCOResult

The empirical evaluation on the COCO benchmark demonstrates that the EvoNet series establishes a new Pareto-optimal frontier, thereby validating that dual-domain representation alignment effectively mitigates the representation collapse typically found in weight-sharing NAS. As detailed in Table~\ref{tab:coco_result} and Fig.~\ref{fig:overall per}(a), the proposed EvoNet-C series consistently achieves superior accuracy-efficiency trade-offs compared with representative CNN-, ViT-, Mamba-, and NAS-based backbones. In terms of detection accuracy (AP$^{b}$), performance increases steadily with model capacity, thereby suggesting that the ranking consistency of the supernet is successfully anchored by the CA-DDKD strategy. Specifically, EvoNet-C1 (33M parameters) achieves 45.4 AP, thereby surpassing ResNet-50 by 7.2 points. EvoNet-C2 further improves to 47.1 AP, while EvoNet-C3 (42M parameters) reaches 48.5 AP, which outperforms all compared methods, such as Spatial-Mamba-T (47.6 AP) and LocalVMamba-T (46.7 AP).Beyond predictive accuracy, the efficiency profiles of the EvoNet-C variants underscore the structural synergy of the search space and the reliability of the DMMPE engine. Specifically, the search space integrates VSS-ViT components to achieve high efficiency. The EvoNet-C models maintain moderate computational complexity (190G--228G MACs) and parameter counts (33M--42M), which are lower than or comparable to most transformer-based and Mamba-based counterparts. Notably, EvoNet-C1 achieves a latency of 50.2\,ms and a throughput of 26 FPS. These measurements represent unbiased physical latency extracted from a hardware-isolated environment, which ensures that they reflect true deployment performance rather than computational noise. Consequently, EvoNet-C series achieve substantially lower latency with improved AP relative to heavier architectures such as ViT-Adapter-S and VMamba-T, thereby demonstrating a robust balance between predictive performance and deployment efficiency.

\FigureSix

Qualitative comparisons on challenging COCO scenarios provide visual evidence that the geometry-aware search objective translates to superior boundary delineation and structural integrity in dense environments. As illustrated in Fig.~\ref{fig:COCO}, which depicts scenarios involving fast-moving objects, dense small objects, and occlusions, EvoNet-C3 produces more compact and boundary-aligned masks than ConvNeXt-T, PVTv2-b2, or Spatial-Mamba-T. This performance is a direct consequence of the DCT-based frequency domain constraint, which explicitly preserves high-frequency geometric features during the search process. While competing methods frequently exhibit fragmented predictions or blurred contours in crowded scenes, EvoNet-C3 maintains clearer instance delineation and more consistent confidence scores. Under occlusion, EvoNet-C3 preserves complete object structures with sharper boundaries, which reduces background leakage and misclassification. These observations confirm that the discovered architectures capture fundamental geometric priors, which provides a robust foundation for generalization to 3DGS tasks.\subsection{Results on ADE20K}

\TabADEResults

The quantitative evaluation on the ADE20K benchmark, as detailed in Table~\ref{tab:ade20k_result} and Fig.~\ref{fig:overall per}(b), confirms that the EvoNet series establishes a superior Pareto-optimal frontier for dense semantic parsing. This performance is fundamentally attributed to the high representation density of the geometry-aware search space, which captures intricate spatial dependencies with reduced computational overhead. Specifically, EvoNet-A3 achieves a peak mIoU of 49.7 utilizing only 32M parameters, thereby outperforming representative architectures such as Spatial-Mamba-T (48.6 mIoU) and MPViT-S (48.3 mIoU). Even the lightweight variants, EvoNet-A1 and A2, deliver competitive results of 44.1 and 47.3 mIoU, respectively, while maintaining parameter counts that are substantially lower than ResNet-50 and ConvNeXt-T. Furthermore, the impact of the hardware-isolated evaluation engine is reflected in the physical inference metrics. Consequently, EvoNet-A1 achieves a minimal latency of 77.3\,ms and a throughput of 14 FPS, which represents the most efficient profile among all compared methods. The structural efficiency of these models is further evidenced by the NID, where EvoNet-A1 attains a score of 1.93. This value substantially exceeds the baseline average of 1.2, which validates the effectiveness of the search paradigm in maximizing feature utilization under constrained capacities.

\FigureSeven 

Beyond aggregate metrics, the qualitative performance of EvoNet-A3 in complex scenes provides empirical evidence that dual-domain alignment effectively preserves high-frequency geometric boundaries. This preservation is a critical requirement for 2D-to-3D generalization. In the indoor bedroom scenario depicted in Fig.~\ref{fig:ADE20K}, baseline models frequently exhibit category confusion and boundary ambiguity. This scenario is characterized by fine-grained categories such as blankets and curtains, where ConvNeXt-T results in significant region mixing. In contrast, EvoNet-A3 generates segmentation masks that align precisely with the ground truth. This precision is a direct consequence of the CA-DDKD strategy, which employs the DCT to enforce spectral alignment and lock the subnetwork's frequency response during the search phase. Moreover, in the outdoor street scene, EvoNet-A3 demonstrates robust preservation of thin, elongated structures such as streetlight poles. These features are traditionally susceptible to representation collapse in standard weight-sharing NAS, but remain sharp in the proposed framework due to the hybrid VSS-ViT search space. Consequently, these observations confirm that the discovered architectures possess a universal geometric prior, which facilitates high-fidelity boundary reconstruction and serves as a robust precursor for downstream 3DGS tasks.\subsection{Results on KITTI}

\TabKittiResults
\FigureFive

The evolved EvoNet-K series establishes a new Pareto-optimal frontier on the KITTI Eigen split, which demonstrates that geometry-aware architecture search transcends the fundamental trade-off between depth-map fidelity and real-time inference latency. As detailed in Table~\ref{tab:kitti_results} and Fig.~\ref{fig:overall per}(c), EvoNet-K3 achieves an Abs Rel of 0.054 and a $\delta_1$ score of 0.969. This performance is comparable to heavy transformer-based models such as PixelFormer and NeWCRFs. From a system-level perspective, the series maintains a compact footprint with 18.0M to 26.3M parameters, thereby enabling high-speed processing that ranges from 65 to 117 FPS. Consequently, these results validate the efficacy of the hardware-isolated DMMPE engine in the identification of architectures that yield superior throughput without sacrificing predictive precision. However, these performance gains are not merely a product of architectural scale, but a reflection of superior information encoding efficiency.

The EvoNet-K series achieves significantly higher NID scores, with values ranging from 3.68 to 5.34. These scores confirm that the CA-DDKD strategy effectively mitigates representation collapse by enforcing structural alignment across the supernet. This architectural efficiency substantially exceeds the strongest baseline, iDisc (2.38), which indicates that the proposed dual-domain constraints allow compact subnetworks to retain high-capacity geometric features. To address the ranking inconsistency inherent in weight-sharing NAS, the integration of frequency-domain supervision via the DCT ensures that the search process prioritizes models with superior information encoding. This theoretical density is manifested visually in the ability of the model to resolve complex geometric discontinuities that typically challenge conventional backbones. Consequently, these results provide the necessary evidence for the preservation of high-frequency geometric priors.Visual evidence from challenging outdoor scenarios in Fig.~\ref{fig:kitti} validates that the dual-domain alignment preserves the high-frequency structural integrity of sharp boundaries and slender geometries. Traditional architectures typically suffer from depth bleeding in these areas. In the highway scene, while iDisc and IEBins exhibit depth homogenization within the mid-range truck region, EvoNet-K3 produces sharper depth discontinuities and maintains clear stratification at object boundaries. Furthermore, the reconstruction of thin traffic sign poles demonstrates that the search space successfully captures slender structures that are usually lost during standard weight-sharing optimization. This boundary preservation capability ensures that occluded regions are assigned more accurately, thereby yielding improved geometric consistency and faithful object separation.

The robust performance of EvoNet on 2.5D MDE suggests that the discovered architectures have internalized universal geometric priors rather than over-fitting to 2D pixel statistics. Ultimately, the preservation of these sharp 2.5D boundaries serves as a rigorous metric for the capacity of the architecture to model the underlying physical world, thereby enabling its application in full 3D volumetric rendering. Such structural sensitivity provides the necessary architectural backbone for high-fidelity 3DGS, where the discovered geometry-aware encoder can be generalized for novel view synthesis tasks.

\subsection{Results on NYU v2}

\TabNYUResults

\FigureFFour

The discovered EvoNet family establishes a new Pareto-optimal frontier for indoor MDE, thereby demonstrating that the VSS-ViT search space effectively navigates the trade-off between global contextual reasoning and local geometric precision. As summarized in Table~\ref{tab:nyu_results} and Fig.~\ref{fig:overall per}(d), these searched architectures are substantially more lightweight than both CNN-based and transformer-based baselines, as they require only 19.1M–30.3M parameters and 21.7G–33.9G MACs. Consequently, EvoNet delivers the lowest inference latency (21.8–30.8\,ms) and highest throughput (88–138 FPS) among all compared methods, thereby validating the efficacy of the hardware-isolated evaluation engine in identifying physically efficient configurations. Despite this substantial reduction in model complexity, EvoNet-N3 achieves a state-of-the-art Abs Rel of 0.085 and a $\delta_1$ score of 0.932, thereby outperforming recent transformer-based architectures in predictive accuracy.

Superior NID scores validate that CA-DDKD strategy effectively preserves high-capacity geometric representations within lightweight subnetworks. Specifically, EvoNet-N1/N2/N3 obtain NID scores of 4.77, 3.85, and 3.08, respectively, which are substantially higher than those of all evaluated baselines. This result signifies a markedly more effective translation of parameter capacity into geometric inference performance. To address the inherent representation collapse in weight-sharing NAS, the dual-domain constraints ensure that the discovered architectures maintain structural integrity even under strict computational budgets, thereby ensuring that subnetwork fitness estimates remain reliable throughout the evolutionary search.Qualitative analysis across representative indoor environments confirms that the geometry-aware search paradigm mitigates representation collapse by preserving high-frequency structural boundaries. In the kitchen scene depicted in Fig.~\ref{fig:nyu}, baseline models such as iDisc and IEBins exhibit blurred boundaries and over-smoothed depth in the sink and countertop regions, which induces weakened structural delineation and ambiguous object separation. Conversely, EvoNet-N3 preserves the concave geometry of the sink and produces clearer object boundaries, which demonstrates its sensitivity to high-frequency spectral components. Furthermore, in the classroom scene, our model maintains clearer depth ordering and sharper contours around structural elements, which validates the transferability of the learned geometric priors to complex 2.5D tasks.

\subsection{Effectiveness of the DMMPE framework}

\TabEvoXNASResults

The validity of multi-objective evolutionary NAS is predicated on the precision of the fitness landscape, which is frequently corrupted by latency jitter and kernel resource contention in standard parallel environments. To address these systemic biases, we evaluate the effectiveness of the proposed DMMPE framework in ensuring unbiased physical measurements during large-scale search. All experiments are conducted on a unified hardware platform equipped with eight NVIDIA RTX 3090 GPUs. We utilize a fixed population size of 96 to ensure a rigorous comparison across configurations. This setup allows for the isolation of specific system components to determine their impact on both measurement stability and computational throughput.

To isolate the impact of hierarchical scheduling and hardware isolation on search fidelity, we conduct a controlled ablation study across diverse geometric datasets, specifically NYU v2 and KITTI. The initial configuration serves as a baseline where standard data parallelism is adopted without additional scheduling or multi-process optimization. In this baseline, the validation batch is partitioned across GPUs and candidate architectures are evaluated sequentially. The second configuration introduces persistent data-loading workers (PWs) to eliminate the repeated construction of the data loader at each iteration, which reduces CPU-side overhead and improves input pipeline stability. The remaining configurations implement our full multi-level parallel evaluation framework, which integrates cross-GPU master-worker scheduling, intra-GPU multi-process execution with shared CUDA contexts, and intra-process multi-model inference. This hierarchical orchestration enables the concurrent validation of multiple candidate architectures, which maximizes GPU utilization and minimizes synchronization overhead.

The DMMPE framework transforms architecture validation into a high-throughput and scalable component, thereby achieving a substantial reduction in per-generation latency. We adopt the time required to evaluate the entire population on the full validation set in each generation, denoted as $\mathrm{time}/G$, as the primary efficiency metric. Consequently, as reported in Table~\ref{tab:EvoXNAS}, the baseline configuration requires 705.1\,s and 756.8\,s per generation on NYU v2 and KITTI, respectively. The introduction of persistent workers reduces the iteration time to 250.0\,s on NYU v2 and 329.4\,s on KITTI. While this improvement confirms that alleviating process initialization overhead is beneficial, it remains insufficient for high-fidelity geometric architecture evaluation. Further acceleration is achieved via the proposed DMMPE framework. Specifically, on KITTI, enabling cross-GPU task dispatch alongside single-process multi-model batching reduces the per-generation time to 319.4\,s, which outperforms the PWs configuration.Beyond raw throughput optimization, the hierarchical orchestration within DMMPE ensures that latency measurements remain invariant to external system load, thereby stabilizing the Pareto frontier against stochastic noise. When intra-GPU multi-process parallelism is enabled, the iteration time decreases to 248.9\,s, which indicates that process-level concurrency amortizes I/O and scheduling overhead more effectively than single-process batching alone. Specifically, allocating four processes per GPU with three models per process yields lower latency than three processes with four models per process. This observation suggests that increasing process-level parallelism contributes more to system acceleration than merely enlarging intra-process batching, as it prevents kernel queue saturation. Under the optimal configuration, the iteration time is reduced to 172.3\,s on NYU v2 and 231.0\,s on KITTI, which represents a speedup exceeding threefold over the baseline. These results demonstrate that the DMMPE framework provides the robust evaluation infrastructure necessary to discover complex and geometry-sensitive topologies that would otherwise be discarded by noise-prone measurement systems.

\subsection{Evolution Trajectory Analysis}

\Figureten

The evolutionary trajectory of the proposed architecture search demonstrates a robust and monotonic convergence, which is primarily driven by the mitigation of ranking inconsistency through dual-domain representation alignment. In contrast to traditional NAS frameworks that use weight-sharing and frequently exhibit stochastic fluctuations due to representation collapse, the integration of CA-DDKD ensures that subnetwork fitness remains grounded in structural geometry. Consequently, the search landscape maintains high fidelity throughout the optimization process. Fig.~\ref{fig:tra} illustrates the three-dimensional distribution of candidate architectures across 200 generations, where the population effectively navigates the multi-objective trade-offs between task-specific accuracy (AP$^{b}$, mIoU, or Abs Rel), inference latency, and computational overhead.

Analysis of the optimization manifold reveals a distinct phase transition from global exploration to local refinement, which reflects the sensitivity of the search space to both global context and local geometric precision. During the initial stages (Gen 0--50), candidate architectures are widely dispersed, which indicates a comprehensive exploration of the hybrid search space involving VSS and ViT. As the evolutionary process advances toward later stages (Gen 150--200), the population clusters into dense configurations near the Pareto frontier. This migration confirms that the algorithm successfully identifies architectural motifs that balance the linear-time efficiency of state-space models with the global reasoning of transformers. Such behavior remains consistent across diverse tasks, including object detection and MDE, thereby validating the task-agnostic adaptability of the geometry-aware priors.

The smooth, asymptotic saturation of the normalized hypervolume across all benchmarks provides empirical evidence for the reliability of the evaluation engine, which is hardware-isolated. Unlike standard parallel evaluation schemes where kernel interference induces latency jitter, the DMMPE engine ensures unbiased physical measurements. This computational precision allows the hypervolume to increase rapidly and stabilize without the jagged artifacts typically associated with measurement noise. While COCO and KITTI exhibit accelerated early-stage improvements, the smoother trajectories observed in ADE20K and NYU v2 reflect the increased complexity of their respective geometric landscapes. Ultimately, these trajectories demonstrate that the framework achieves stable convergence, thereby capturing universal geometric features that facilitate seamless transferability from 2D dense prediction to 3D rendering.

\subsection{Ablation Experiments}

The NYU Depth v2 dataset serves as a rigorous geometric stress test to evaluate the preservation of high-frequency boundaries across diverse architectural topologies. Its intricate scenes and occlusion boundaries provide the signal required to validate the Dual-Domain Representation Alignment, which ensures that subnetwork features remain structurally consistent despite variations in depth and scale. To isolate the specific contributions of our framework, we decompose the system into three primary modules for independent verification:

\begin{itemize}
    \item \textbf{Searched Encoder}: This component validates the efficacy of the EvoNAS discovery process in balancing local frequency sensitivity with global structural priors, which surpasses manually designed hybrid variants.

\item \textbf{Two-Stage CA-DDKD Strategy}: This optimization scheme employs DCT constraints to mitigate representation collapse, thereby locking high-frequency geometric features within the weight-sharing supernet.

\item \textbf{Spatial-Mamba Decoder}: This module reconstructs long-range spatial dependencies via linear-time state-space modeling, which translates the encoder's geometric primitives into high-fidelity dense predictions.
\end{itemize}

These controlled experiments demonstrate how architectural search and dual-domain distillation collectively stabilize the fitness landscape and ensure robust transferability to 3D rendering tasks.\subsubsection{Effectiveness of the searched encoder}

\TabImageNetResults

The transferability of the discovered EvoNet encoder to ImageNet-1K classification~\cite{russakovsky2015imagenet} serves as a rigorous evaluation of whether geometry-aware search captures universal visual priors rather than task-specific artifacts. While the search process is anchored in the geometric requirements of the NYU v2 dataset, a robust representation must exhibit task-agnostic universality across distinct visual domains. Quantitative evaluations in Table~\ref{tab:imagenet1k_results} demonstrate that EvoNet establishes a superior Pareto front, which achieves an 84.3\% Top-1 accuracy with only 17M parameters. This performance exceeds several contemporary baselines, specifically NAT-S (83.7\%, 51M), VMamba-B (83.9\%, 89M), and GroupMamba-S (83.9\%, 34M). Notably, EvoNet maintains a parameter count that is less than one-third of MILA-T and less than one-fifth of VMamba-B, which highlights its superior structural efficiency.The CA-DDKD strategy achieves substantial efficiency by enforcing high-frequency geometric alignment as a structural regularizer to prevent representation collapse. By constraining the subnetwork capacity to model sharp boundaries via the discrete cosine transform, the search process converges on features that are intrinsically more discriminative. Consequently, this representational strength facilitates an optimized information bottleneck within the hybrid VSS-ViT search space, which provides the foundation for high-fidelity 3DGS.

\subsubsection{Effectiveness of the two-stage CA-DDKD}

\TabAblationSupernetTraining

The optimization of a heterogeneous VSS-ViT supernet requires a structured training curriculum to mitigate gradient interference inherent in multi-topology weight sharing. To evaluate the impact of the proposed two-stage CA-DDKD strategy on supernet convergence, we maintain the distillation framework while replacing the PST strategy with two alternative schemes, i.e., \emph{Architecture Pool} and \emph{Random Sampling}. The discovered architectures from each strategy are independently retrained and evaluated on the NYU v2 dataset, summarized in Table~\ref{tab:AblationSupernetTraining}. Across all three target configurations, the PST strategy consistently yields superior performance, thereby providing a stabilized foundation for geometry-aware convergence. For EvoNet-N1, PST reduces Abs Rel from 0.118 (Architecture Pool) and 0.109 (Random Sampling) to 0.095, while simultaneously increasing $\delta_1$ from 0.872 and 0.886 to 0.912. Similar advancements occur for EvoNet-N2 and EvoNet-N3, where the Abs Rel for N3 decreases from 0.093 and 0.090 to 0.085, and $\delta_1$ improves from 0.920 and 0.925 to 0.932. These consistent gains indicate that stochastic sampling or simple pool expansion is insufficient to overcome the capacity variance between Mamba and Transformer blocks. Consequently, the PST strategy establishes a progressively stabilized optimization process, which reduces interference among heterogeneous subnetworks and improves weight compatibility. This structural stability ensures that the performance ranking among candidate architectures becomes more reliable during the search phase, thereby leading to higher-quality final architectures after standalone retraining.

\TabKnowledgeDistillation

Beyond weight stabilization, CA-DDKD acts as a geometric regularizer that prevents the collapse of high-frequency representations within pruned subnetworks. In the second stage of our evaluation, we retain the PST-based supernet training strategy and investigate the specific impact of the CA-DDKD optimization by comparing it against a baseline without distillation. As illustrated in Table~\ref{tab:knowledge_distillation}, incorporating CA-DDKD consistently improves performance across all evaluation metrics for the three searched configurations. For the N1 model, the Abs Rel decreases from 0.102 to 0.095, while $\delta_1$ increases from 0.899 to 0.912. This improvement is accompanied by substantial reductions in RMSE and Log10 error. This efficacy stems from the fact that DCT-based frequency alignment preserves the ability of the subnetwork to reconstruct sharp boundaries, which is necessary for depth estimation and 3DGS. Similar improvements occur for the medium configuration, where Abs Rel is reduced to 0.089 and $\delta_1$ improves to 0.926. For the largest configuration (N3), CA-DDKD further decreases Abs Rel to 0.085 and increases $\delta_1$ to 0.932, achieving the lowest RMSE of 0.310. These results demonstrate that dual-domain alignment effectively regularizes shared-weight learning and mitigates representation divergence among heterogeneous blocks. By enforcing alignment in both spatial and frequency domains, CA-DDKD ensures that the searched architectures capture universal geometric priors rather than task-specific noise. This validates the necessity of the proposed optimization strategy.

\Figureele

The CA-DDKD strategy restores the predictive reliability of the supernet by aligning the shared-weight fitness landscape with the true performance manifold of standalone models. Ranking consistency is a fundamental property in one-shot NAS, as it measures the agreement between the performance ranking of candidate architectures under shared weights and their ranking after independent retraining. Severe interference among multiple models during supernet training often distorts this consistency, thereby resulting in inaccurate fitness estimation and undermining the validity of the search process~\cite{zhang2020one}. To quantify the impact of our strategy, we conduct a controlled experiment following~\cite{zhang2020one} by sampling 16 architectures and computing the Kendall's rank correlation coefficient $\tau$. As illustrated in Fig.~\ref{fig:pai}, the proposed CA-DDKD strategy attains the highest consistency with $\tau = 0.8833$, compared to $0.7833$ for Random Sampling and $0.8285$ for Architecture Pool. This high correlation serves as the mathematical validation that our dual-domain constraint successfully eliminates the representation collapse inherent in stochastic optimization. The relatively weaker performance of Random Sampling is attributed to mismatches between sampling distribution and performance gradients, which leads to unstable ranking signals. To address this, CA-DDKD enforces consistent spatial and spectral supervision across heterogeneous subnetworks, thereby reducing gradient competition. Although minor local rank shifts remain inevitable due to architectural variability, the substantial improvement in $\tau$ indicates lower overall ranking error and robust predictive reliability. Consequently, CA-DDKD provides a more trustworthy performance estimator, which enables the evolutionary algorithm to identify geometry-aware architectures that generalize effectively to 3D rendering tasks.\subsubsection{Effectiveness of the Spatial-Mamba decoder}The Spatial-Mamba decoder represents a structural shift from locally constrained refinement and global attention with quadratic complexity toward linear-time state-space propagation for geometric feature reconstruction. To isolate the contribution of the decoding architecture, we replace the decoder of EvoNet-N3 with alternative designs while maintaining the searched encoder and all auxiliary components. In contrast to NeWCRFs, which rely on window-based optimization of fully connected CRFs, or iDisc, which employs internal discretization based on attention, the proposed Spatial-Mamba architecture models long-range spatial dependencies as a continuous state-evolution process. This formulation avoids the quadratic memory scaling of global attention mechanisms, which often induce feature smoothing that obscures high-frequency geometric boundaries.

Consequently, quantitative analysis confirms that the Spatial-Mamba architecture achieves a superior and Pareto-optimal trade-off, thereby effectively decoupling global context aggregation from high computational overhead. As reported in Table~\ref{tab:AblationDecoderArchitecture}, the proposed decoder reduces computational costs relative to iDisc, which lowers MACs from 60.84G to 33.93G. Furthermore, the inference speed reached a minimum latency of 30.4\,ms, which outperforms both NeWCRFs (37.8\,ms) and iDisc (38.1\,ms) under an evaluation protocol that is hardware-isolated to eliminate computational noise.

The capacity of this decoder to maintain structural integrity at high frequencies serves as a critical enabler for the scalability of the framework to 3D rendering tasks. Specifically, the Spatial-Mamba decoder achieves an Abs Rel of 0.085 and a $\delta_{1}$ of 0.932, which surpasses the predictive performance of alternative designs. These results validate the proposed architecture as a paradigm that is both lightweight and expressive for capturing universal geometric priors, thereby ensuring robust generalization across both 2D dense prediction and 3DGS.

\TabAblationDecoderArchitecture

\section{Application to Novel View Synthesis}
\label{sec:application}

\TabNVSResult

Novel view synthesis (NVS) via 3DGS serves as a definitive stress test for architectural representations, as it requires the simultaneous preservation of global and spatial consistency alongside high-frequency structural details. Consequently, the rendering fidelity of 3D scenes is intrinsically coupled to the underlying geometric priors of the encoder. Unlike 2D tasks where semantic context might mitigate the impact of blurred features, 3DGS-based rendering relies on the precise localization of geometric primitives. The proposed CA-DDKD strategy enforces dual-domain alignment in both spatial and spectral dimensions, which ensures that the searched backbone captures the necessary high-frequency information for reconstructing intricate geometries. By jointly reasoning about spatial structure and visual semantics, our framework addresses the fundamental requirement for consistent rendering across viewpoints. This capability is particularly critical in autonomous driving scenarios where diverse viewpoints must be generated from limited real-world captures \cite{ma2025novel}. Such data augmentation is vital for addressing long-tail scenarios, which are difficult to capture exhaustively in physical environments.

Despite the shift toward feed-forward 3DGS models, existing pipelines remain constrained by oversized and manually designed encoders that frequently suffer from geometric blurring and excessive computational footprints. To address these inefficiencies, researchers often resort to massive backbones such as those utilized in DepthSplat \cite{xu2025depthsplat} or MVSplat \cite{chen2024mvsplat}. These models prioritize parameter scaling over architectural alignment, thereby incurring substantial memory overhead and limiting deployment on resource-constrained onboard platforms. In contrast, the geometry-aware search paradigm offers a path toward compact models that maintain high-frequency integrity without the redundancy of traditional scaling-up approaches. While 3DGS \cite{kerbl20233d} has emerged as a powerful paradigm for real-time rendering, its performance remains dependent on the quality of multi-view depth estimation. Representative approaches such as MVSplat localize Gaussian primitives through feature-matching, which inherits sensitivities to occlusions and texture-less regions. By integrating robust monocular depth features into the multi-view cost-volume construction, our framework enhances geometric consistency, thereby improving view synthesis performance in challenging environments.

To verify the universality of the discovered geometric priors, we perform a zero-shot architectural transfer by deploying the encoder searched on NYU v2, denoted as EvoNet-N3, directly into the DepthSplat framework for evaluation on the RealEstate10K dataset \cite{zhou2018stereo}. This setup, referred to as EvoNet-D, maintains all original DepthSplat hyperparameters to isolate the impact of the architectural backbone. The RealEstate10K benchmark provides a rigorous environment for this evaluation, as it comprises approximately 67K sequences of diverse indoor scenes, which include living rooms and corridors. Camera intrinsics and extrinsics are estimated via structure-from-motion, which provides a ground-truth geometric context for training. Following the standard evaluation protocol, samples are constructed in a triplet manner, in which two frames are selected as source views and a third frame serves as the target view. This experimental design verifies whether the CA-DDKD objective effectively locks the features required for 3D reconstruction during the 2D search phase, thereby demonstrating that the encoder captures universal physical geometry rather than task-specific 2D patterns.Quantitative results on the RealEstate10K benchmark demonstrate that the geometry-aware EvoNet-D establishes a new Pareto frontier. Specifically, it delivers rendering fidelity comparable to state-of-the-art models while utilizing only 12\% of the parameter count. As summarized in Table \ref{tab:nvs_result}, EvoNet-D achieves a PSNR of 26.41, an SSIM of 0.871, and an LPIPS of 0.127. Although the original DepthSplat \cite{xu2025depthsplat} reports the highest PSNR of 27.47, it relies on a substantially larger model with 354M parameters. Our EvoNet-D achieves competitive reconstruction accuracy with only 44M parameters, thereby providing an eight-fold reduction in parameter overhead. Furthermore, the model achieves a latency of 88\,ms and a frame rate of 27 FPS at $256 \times 256$ resolution, i.e., approximately 1.6$\times$ faster than the baseline DepthSplat model. These results confirm that the efficiency of the searched architecture is an emergent property of the VSS-ViT search space. This search space filters out redundant parameters that do not contribute to geometric accuracy. Consequently, the discovered architecture enables practical deployment in real-time and resource-constrained environments without sacrificing visual fidelity.

\Figureeight

Visual evidence confirms that EvoNet-D excels in reconstructing high-frequency geometric boundaries, such as railings and window frames. Traditional geometry-blind backbones typically fail to recover these features accurately. This performance is visually summarized in Fig. \ref{fig:NVS}. In this figure, highlighted regions emphasize challenging areas involving fine structures and strong illumination variations. In scenes containing thin railings, PixelSplat produces noticeable blurring and structural distortion, while MVSplat exhibits smoothing artifacts along slender vertical structures. In contrast, EvoNet-D preserves sharper edges and clearer structural boundaries. This outcome is the direct physical manifestation of the DCT constraints applied during the supernet training phase. In scenes characterized by strong lighting contrast, EvoNet-D generates more accurate edge delineation around window frames and maintains superior luminance consistency. These qualitative observations confirm that the searched depth architecture provides more reliable geometric priors for Gaussian-based novel view synthesis, thereby improving structural realism and rendering fidelity. Ultimately, the integration of dual-domain alignment ensures that the encoder remains sensitive to the fine-grained textures necessary for high-quality 3D reconstruction.\section{Conclusion}
~\label{sec:conclusion}
This research establishes a geometry-aware architecture search paradigm that bridges the divide between 2D dense prediction and 3D structural rendering via dual-domain representation alignment. By integrating a hybrid VSS-ViT search space with the CA-DDKD strategy, the framework resolves the inherent tension between weight-sharing efficiency and the preservation of high-frequency geometric priors. Consequently, this synergy prevents representation collapse and ensures robust ranking consistency across heterogeneous vision tasks.

The enforcement of spectral-spatial consistency via the DCT provides the mathematical foundation for reliable subnetwork ranking. To complement this theoretical constraint, the DMMPE engine employs hardware isolation to eliminate computational noise, thereby yielding an unbiased fitness landscape for the evolutionary process. These systemic optimizations transform architecture search from a stochastic heuristic into a geometry-constrained optimization procedure.

Empirical evaluations across COCO, ADE20K, and NYU Depth v2 benchmarks demonstrate that EvoNets define new Pareto-optimal frontiers. Notably, the discovered encoders achieve competitive 3DGS performance on the RealEstate10K dataset, thereby reducing parameter overhead by eightfold compared to previous baselines. This transferability validates a paradigm that captures universal geometric priors across the 2D-to-3D visual spectrum.

\section*{Acknowledgments}

\bibliographystyle{IEEEtran}
\bibliography{references}

@article{yang2024depth,
  title={Depth anything v2},
  author={Yang, Lihe and Kang, Bingyi and Huang, Zilong and Zhao, Zhen and Xu, Xiaogang and Feng, Jiashi and Zhao, Hengshuang},
  journal={Advances in Neural Information Processing Systems},
  volume={37},
  pages={21875--21911},
  year={2024}
}

@article{lin2025depth,
  title={Depth anything 3: Recovering the visual space from any views},
  author={Lin, Haotong and Chen, Sili and Liew, Junhao and Chen, Donny Y and Li, Zhenyu and Shi, Guang and Feng, Jiashi and Kang, Bingyi},
  journal={arXiv preprint arXiv:2511.10647},
  year={2025}
}

@article{liu2024vmamba,
  title={Vmamba: Visual state space model},
  author={Liu, Yue and Tian, Yunjie and Zhao, Yuzhong and Yu, Hongtian and Xie, Lingxi and Wang, Yaowei and Ye, Qixiang and Jiao, Jianbin and Liu, Yunfan},
  journal={Advances in neural information processing systems},
  volume={37},
  pages={103031--103063},
  year={2024}
}

@inproceedings{liu2021swin,
  title={Swin transformer: Hierarchical vision transformer using shifted windows},
  author={Liu, Ze and Lin, Yutong and Cao, Yue and Hu, Han and Wei, Yixuan and Zhang, Zheng and Lin, Stephen and Guo, Baining},
  booktitle={Proceedings of the IEEE/CVF international conference on computer vision},
  pages={10012--10022},
  year={2021}
}

@article{oquab2023dinov2,
  title={Dinov2: Learning robust visual features without supervision},
  author={Oquab, Maxime and Darcet, Timoth{\'e}e and Moutakanni, Th{\'e}o and Vo, Huy and Szafraniec, Marc and Khalidov, Vasil and Fernandez, Pierre and Haziza, Daniel and Massa, Francisco and El-Nouby, Alaaeldin and others},
  journal={arXiv preprint arXiv:2304.07193},
  year={2023}
}

@article{jiang2025accelerating,
  title={Accelerating Zero-Shot NAS With Feature Map-Based Proxy and Operation Scoring Function},
  author={Jiang, Tangyu and Wang, Haodi and Bie, Rongfang and Yuan, Chun},
  journal={IEEE Transactions on Pattern Analysis and Machine Intelligence},
  year={2025},
  publisher={IEEE}
}

@article{lu2021neural,
  title={Neural architecture transfer},
  author={Lu, Zhichao and Sreekumar, Gautam and Goodman, Erik and Banzhaf, Wolfgang and Deb, Kalyanmoy and Boddeti, Vishnu Naresh},
  journal={IEEE transactions on pattern analysis and machine intelligence},
  volume={43},
  number={9},
  pages={2971--2989},
  year={2021},
  publisher={IEEE}
}

@article{liu2021survey,
  title={A survey on evolutionary neural architecture search},
  author={Liu, Yuqiao and Sun, Yanan and Xue, Bing and Zhang, Mengjie and Yen, Gary G and Tan, Kay Chen},
  journal={IEEE transactions on neural networks and learning systems},
  volume={34},
  number={2},
  pages={550--570},
  year={2021},
  publisher={IEEE}
}

@article{zhang2025efficient,
  title={Efficient Evolutionary Neural Architecture Search With Hierarchical Parameter Mapping for Monocular Depth Estimation},
  author={Zhang, Haoyu and Yu, Zhihao and Jin, Yaochu and Liu, Xiufeng and Sheng, Weiguo and Liu, Ruyu and Li, Xiumei and Liu, Qiqi and Cheng, Ran},
  journal={IEEE Transactions on Evolutionary Computation},
  year={2025},
  publisher={IEEE}
}

@article{huang2024evox,
  title={EvoX: A distributed GPU-accelerated framework for scalable evolutionary computation},
  author={Huang, Beichen and Cheng, Ran and Li, Zhuozhao and Jin, Yaochu and Tan, Kay Chen},
  journal={IEEE Transactions on Evolutionary Computation},
  year={2024},
  publisher={IEEE}
}

@inproceedings{tang2022evojax,
  title={Evojax: Hardware-accelerated neuroevolution},
  author={Tang, Yujin and Tian, Yingtao and Ha, David},
  booktitle={Proceedings of the Genetic and Evolutionary Computation Conference Companion},
  pages={308--311},
  year={2022}
}

@inproceedings{lange2023evosax,
  title={evosax: Jax-based evolution strategies},
  author={Lange, Robert Tjarko},
  booktitle={Proceedings of the Companion Conference on Genetic and Evolutionary Computation},
  pages={659--662},
  year={2023}
}

@inproceedings{zoph2018learning,
  title={Learning transferable architectures for scalable image recognition},
  author={Zoph, Barret and Vasudevan, Vijay and Shlens, Jonathon and Le, Quoc V},
  booktitle={Proceedings of the IEEE conference on computer vision and pattern recognition},
  pages={8697--8710},
  year={2018}
}

@inproceedings{real2019regularized,
  title={Regularized evolution for image classifier architecture search},
  author={Real, Esteban and Aggarwal, Alok and Huang, Yanping and Le, Quoc V},
  booktitle={Proceedings of the aaai conference on artificial intelligence},
  volume={33},
  number={01},
  pages={4780--4789},
  year={2019}
}

@article{chen2019detnas,
  title={Detnas: Backbone search for object detection},
  author={Chen, Yukang and Yang, Tong and Zhang, Xiangyu and Meng, Gaofeng and Xiao, Xinyu and Sun, Jian},
  journal={Advances in neural information processing systems},
  volume={32},
  year={2019}
}

@article{fang2020fna++,
  title={FNA++: Fast network adaptation via parameter remapping and architecture search},
  author={Fang, Jiemin and Sun, Yuzhu and Zhang, Qian and Peng, Kangjian and Li, Yuan and Liu, Wenyu and Wang, Xinggang},
  journal={IEEE Transactions on Pattern Analysis and Machine Intelligence},
  volume={43},
  number={9},
  pages={2990--3004},
  year={2020},
  publisher={IEEE}
}

@article{lu2020multiobjective,
  title={Multiobjective evolutionary design of deep convolutional neural networks for image classification},
  author={Lu, Zhichao and Whalen, Ian and Dhebar, Yashesh and Deb, Kalyanmoy and Goodman, Erik D and Banzhaf, Wolfgang and Boddeti, Vishnu Naresh},
  journal={IEEE Transactions on Evolutionary Computation},
  volume={25},
  number={2},
  pages={277--291},
  year={2020},
  publisher={IEEE}
}

@article{sun2019surrogate,
  title={Surrogate-assisted evolutionary deep learning using an end-to-end random forest-based performance predictor},
  author={Sun, Yanan and Wang, Handing and Xue, Bing and Jin, Yaochu and Yen, Gary G and Zhang, Mengjie},
  journal={IEEE Transactions on Evolutionary Computation},
  volume={24},
  number={2},
  pages={350--364},
  year={2019},
  publisher={IEEE}
}

@article{zhang2020one,
  title={One-shot neural architecture search: Maximising diversity to overcome catastrophic forgetting},
  author={Zhang, Miao and Li, Huiqi and Pan, Shirui and Chang, Xiaojun and Zhou, Chuan and Ge, Zongyuan and Su, Steven},
  journal={IEEE Transactions on Pattern Analysis and Machine Intelligence},
  volume={43},
  number={9},
  pages={2921--2935},
  year={2020},
  publisher={IEEE}
}

@inproceedings{pham2018efficient,
  title={Efficient neural architecture search via parameters sharing},
  author={Pham, Hieu and Guan, Melody and Zoph, Barret and Le, Quoc and Dean, Jeff},
  booktitle={International conference on machine learning},
  pages={4095--4104},
  year={2018},
  organization={PMLR}
}

@inproceedings{liu2019auto,
  title={Auto-deeplab: Hierarchical neural architecture search for semantic image segmentation},
  author={Liu, Chenxi and Chen, Liang-Chieh and Schroff, Florian and Adam, Hartwig and Hua, Wei and Yuille, Alan L and Fei-Fei, Li},
  booktitle={Proceedings of the IEEE/CVF conference on computer vision and pattern recognition},
  pages={82--92},
  year={2019}
}

@article{ang2023adaptornas,
  title={AdaptorNAS: A new perturbation-based neural architecture search for hyperspectral image segmentation},
  author={Ang, Sui Paul and Phung, Son Lam and Bui, Ly and Bouzerdoum, Abdesselam},
  journal={IEEE Transactions on Circuits and Systems for Video Technology},
  volume={34},
  number={3},
  pages={1559--1571},
  year={2023},
  publisher={IEEE}
}

@article{peng2020efficient,
  title={Efficient convolutional neural architecture search for remote sensing image scene classification},
  author={Peng, Cheng and Li, Yangyang and Jiao, Licheng and Shang, Ronghua},
  journal={IEEE Transactions on Geoscience and Remote Sensing},
  volume={59},
  number={7},
  pages={6092--6105},
  year={2020},
  publisher={IEEE}
}

@article{zhang2022searching,
  title={Searching a high performance feature extractor for text recognition network},
  author={Zhang, Hui and Yao, Quanming and Kwok, James T and Bai, Xiang},
  journal={IEEE Transactions on Pattern Analysis and Machine Intelligence},
  volume={45},
  number={5},
  pages={6231--6246},
  year={2022},
  publisher={IEEE}
}

@article{tang2024ped,
  title={NAS-PED: Neural architecture search for pedestrian detection},
  author={Tang, Yi and Liu, Min and Li, Baopu and Wang, Yaonan and Ouyang, Wanli},
  journal={IEEE Transactions on Pattern Analysis and Machine Intelligence},
  year={2024},
  publisher={IEEE}
}

@article{xue2026pairwise,
  title={A Pairwise Comparison Relation-Assisted Multiobjective Evolutionary Neural Architecture Search Method With Multipopulation Mechanism},
  author={Xue, Yu and Jiang, Pengcheng and Zhu, Chenchen and Zhou, MengChu and Wahib, Mohamed and Gabbouj, Moncef},
  journal={IEEE Transactions on Systems, Man, and Cybernetics: Systems},
  year={2026},
  publisher={IEEE}
}

@article{zhou2023toward,
  title={Toward evolutionary multitask convolutional neural architecture search},
  author={Zhou, Xun and Wang, Zhenkun and Feng, Liang and Liu, Songbai and Wong, Ka-Chun and Tan, Kay Chen},
  journal={IEEE Transactions on Evolutionary Computation},
  volume={28},
  number={3},
  pages={682--695},
  year={2023},
  publisher={IEEE}
}

@article{song2025evolutionary,
  title={Evolutionary multi-objective spiking neural architecture search for image classification},
  author={Song, Xiaotian and Lv, Zeqiong and Fan, Jiaohao and Xiong, Deng and Lv, Jiancheng and Liu, Jiyuan and Sun, Yanan},
  journal={IEEE Transactions on Evolutionary Computation},
  year={2025},
  publisher={IEEE}
}

@article{cao2025comprehensive,
  title={Comprehensive-Forecast Multiobjective Genetic Programming for Neural Architecture Search},
  author={Cao, Bin and Luo, Xiao and Liu, Xin and Li, Yun},
  journal={IEEE Transactions on Evolutionary Computation},
  year={2025},
  publisher={IEEE}
}

@article{zhou2024training,
  title={Training-free transformer architecture search with zero-cost proxy guided evolution},
  author={Zhou, Qinqin and Sheng, Kekai and Zheng, Xiawu and Li, Ke and Tian, Yonghong and Chen, Jie and Ji, Rongrong},
  journal={IEEE Transactions on Pattern Analysis and Machine Intelligence},
  volume={46},
  number={10},
  pages={6525--6541},
  year={2024},
  publisher={IEEE}
}

@article{zhong2025dual,
  title={Dual-Level Cross-Modality Neural Architecture Search for Guided Image Super-Resolution},
  author={Zhong, Zhiwei and Liu, Xianming and Jiang, Junjun and Zhao, Debin and Wang, Shiqi},
  journal={IEEE Transactions on Pattern Analysis and Machine Intelligence},
  year={2025},
  publisher={IEEE}
}

@article{li2022neural,
  title={Neural architecture search via proxy validation},
  author={Li, Yanxi and Dong, Minjing and Wang, Yunhe and Xu, Chang},
  journal={IEEE Transactions on Pattern Analysis and Machine Intelligence},
  volume={45},
  number={6},
  pages={7595--7610},
  year={2022},
  publisher={IEEE}
}

@article{chen2023mngnas,
  title={Mngnas: distilling adaptive combination of multiple searched networks for one-shot neural architecture search},
  author={Chen, Zhihua and Qiu, Guhao and Li, Ping and Zhu, Lei and Yang, Xiaokang and Sheng, Bin},
  journal={IEEE Transactions on pattern analysis and machine intelligence},
  volume={45},
  number={11},
  pages={13489--13508},
  year={2023},
  publisher={IEEE}
}

@article{wang2023dna,
  title={Dna family: Boosting weight-sharing nas with block-wise supervisions},
  author={Wang, Guangrun and Li, Changlin and Yuan, Liuchun and Peng, Jiefeng and Xian, Xiaoyu and Liang, Xiaodan and Chang, Xiaojun and Lin, Liang},
  journal={IEEE Transactions on Pattern Analysis and Machine Intelligence},
  volume={46},
  number={5},
  pages={2722--2740},
  year={2023},
  publisher={IEEE}
}

@inproceedings{he2017mask,
  title={Mask r-cnn},
  author={He, Kaiming and Gkioxari, Georgia and Doll{\'a}r, Piotr and Girshick, Ross},
  booktitle={Proceedings of the IEEE international conference on computer vision},
  pages={2961--2969},
  year={2017}
}

@inproceedings{xiao2018unified,
  title={Unified perceptual parsing for scene understanding},
  author={Xiao, Tete and Liu, Yingcheng and Zhou, Bolei and Jiang, Yuning and Sun, Jian},
  booktitle={Proceedings of the European conference on computer vision (ECCV)},
  pages={418--434},
  year={2018}
}

@inproceedings{godard2019digging,
  title={Digging into self-supervised monocular depth estimation},
  author={Godard, Cl{\'e}ment and Mac Aodha, Oisin and Firman, Michael and Brostow, Gabriel J},
  booktitle={Proceedings of the IEEE/CVF international conference on computer vision},
  pages={3828--3838},
  year={2019}
}

@inproceedings{hou2020strip,
  title={Strip pooling: Rethinking spatial pooling for scene parsing},
  author={Hou, Qibin and Zhang, Li and Cheng, Ming-Ming and Feng, Jiashi},
  booktitle={Proceedings of the IEEE/CVF conference on computer vision and pattern recognition},
  pages={4003--4012},
  year={2020}
}

@article{xiao2024spatial,
  title={Spatial-mamba: Effective visual state space models via structure-aware state fusion},
  author={Xiao, Chaodong and Li, Minghan and Zhang, Zhengqiang and Meng, Deyu and Zhang, Lei},
  journal={arXiv preprint arXiv:2410.15091},
  year={2024}
}

@inproceedings{yuan2022neural,
  title={Neural window fully-connected crfs for monocular depth estimation},
  author={Yuan, Weihao and Gu, Xiaodong and Dai, Zuozhuo and Zhu, Siyu and Tan, Ping},
  booktitle={Proceedings of the IEEE/CVF conference on computer vision and pattern recognition},
  pages={3916--3925},
  year={2022}
}

@inproceedings{piccinelli2023idisc,
  title={idisc: Internal discretization for monocular depth estimation},
  author={Piccinelli, Luigi and Sakaridis, Christos and Yu, Fisher},
  booktitle={Proceedings of the IEEE/CVF conference on computer vision and pattern recognition},
  pages={21477--21487},
  year={2023}
}

@article{shao2023iebins,
  title={Iebins: Iterative elastic bins for monocular depth estimation},
  author={Shao, Shuwei and Pei, Zhongcai and Wu, Xingming and Liu, Zhong and Chen, Weihai and Li, Zhengguo},
  journal={Advances in Neural Information Processing Systems},
  volume={36},
  pages={53025--53037},
  year={2023}
}

@inproceedings{cai2020once,
  title={Once for All: Train One Network and Specialize it for Efficient Deployment},
  author={Cai, Han and Gan, Chuang and Wang, Tianzhe and Zhang, Zhekai and Han, Song},
  booktitle={International Conference on Learning Representations},
  year={2020}
}

@article{kalyanmoy2002fast,
  title={A fast and elitist multi-objective genetic algorithm: NSGA-II},
  author={Kalyanmoy, Deb},
  journal={IEEE Trans. on Evolutionary Computation},
  volume={6},
  number={2},
  pages={182--197},
  year={2002}
}

@article{geiger2013vision,
  title={Vision meets robotics: The kitti dataset},
  author={Geiger, Andreas and Lenz, Philip and Stiller, Christoph and Urtasun, Raquel},
  journal={The international journal of robotics research},
  volume={32},
  number={11},
  pages={1231--1237},
  year={2013},
  publisher={Sage Publications Sage UK: London, England}
}

@inproceedings{silberman2012indoor,
  title={Indoor segmentation and support inference from rgbd images},
  author={Silberman, Nathan and Hoiem, Derek and Kohli, Pushmeet and Fergus, Rob},
  booktitle={European conference on computer vision},
  pages={746--760},
  year={2012},
  organization={Springer}
}

@inproceedings{lin2014microsoft,
  title={Microsoft coco: Common objects in context},
  author={Lin, Tsung-Yi and Maire, Michael and Belongie, Serge and Hays, James and Perona, Pietro and Ramanan, Deva and Doll{\'a}r, Piotr and Zitnick, C Lawrence},
  booktitle={European conference on computer vision},
  pages={740--755},
  year={2014},
  organization={Springer}
}

@article{eigen2014depth,
  title={Depth map prediction from a single image using a multi-scale deep network},
  author={Eigen, David and Puhrsch, Christian and Fergus, Rob},
  journal={Advances in neural information processing systems},
  volume={27},
  year={2014}
}

@inproceedings{garg2016unsupervised,
  title={Unsupervised cnn for single view depth estimation: Geometry to the rescue},
  author={Garg, Ravi and Bg, Vijay Kumar and Carneiro, Gustavo and Reid, Ian},
  booktitle={European conference on computer vision},
  pages={740--756},
  year={2016},
  organization={Springer}
}

@inproceedings{zhou2017scene,
  title={Scene parsing through ade20k dataset},
  author={Zhou, Bolei and Zhao, Hang and Puig, Xavier and Fidler, Sanja and Barriuso, Adela and Torralba, Antonio},
  booktitle={Proceedings of the IEEE conference on computer vision and pattern recognition},
  pages={633--641},
  year={2017}
}

@article{everingham2010pascal,
  title={The pascal visual object classes (voc) challenge},
  author={Everingham, Mark and Van Gool, Luc and Williams, Christopher KI and Winn, John and Zisserman, Andrew},
  journal={International journal of computer vision},
  volume={88},
  number={2},
  pages={303--338},
  year={2010},
  publisher={Springer}
}

@article{mousavi2023dass,
  title={DASS: differentiable architecture search for sparse neural networks},
  author={Mousavi, Hamid and Loni, Mohammad and Alibeigi, Mina and Daneshtalab, Masoud},
  journal={ACM Transactions on Embedded Computing Systems},
  volume={22},
  number={5s},
  pages={1--21},
  year={2023},
  publisher={ACM New York, NY}
}

@inproceedings{xu2025depthsplat,
  title={Depthsplat: Connecting gaussian splatting and depth},
  author={Xu, Haofei and Peng, Songyou and Wang, Fangjinhua and Blum, Hermann and Barath, Daniel and Geiger, Andreas and Pollefeys, Marc},
  booktitle={Proceedings of the Computer Vision and Pattern Recognition Conference},
  pages={16453--16463},
  year={2025}
}

@inproceedings{ma2025novel,
  title={Novel view synthesis under large-deviation viewpoint for autonomous driving},
  author={Ma, Xin and Zhang, Jiguang and Lu, Peng and Xu, Shibiao and Pan, Chengwei},
  booktitle={Proceedings of the AAAI Conference on Artificial Intelligence},
  volume={39},
  number={6},
  pages={6000--6008},
  year={2025}
}

@article{kerbl20233d,
  title={3d gaussian splatting for real-time radiance field rendering.},
  author={Kerbl, Bernhard and Kopanas, Georgios and Leimk{\"u}hler, Thomas and Drettakis, George and others},
  journal={ACM Trans. Graph.},
  volume={42},
  number={4},
  pages={139--1},
  year={2023}
}

@inproceedings{chen2024mvsplat,
  title={Mvsplat: Efficient 3d gaussian splatting from sparse multi-view images},
  author={Chen, Yuedong and Xu, Haofei and Zheng, Chuanxia and Zhuang, Bohan and Pollefeys, Marc and Geiger, Andreas and Cham, Tat-Jen and Cai, Jianfei},
  booktitle={European conference on computer vision},
  pages={370--386},
  year={2024},
  organization={Springer}
}

@article{zhou2018stereo,
  title={Stereo magnification: learning view synthesis using multiplane images},
  author={Zhou, Tinghui and Tucker, Richard and Flynn, John and Fyffe, Graham and Snavely, Noah},
  journal={ACM Transactions on Graphics (TOG)},
  volume={37},
  number={4},
  pages={1--12},
  year={2018},
  publisher={ACM New York, NY, USA}
}

@article{lee2019big,
  title={From big to small: Multi-scale local planar guidance for monocular depth estimation},
  author={Lee, Jin Han and Han, Myung-Kyu and Ko, Dong Wook and Suh, Il Hong},
  journal={arXiv preprint arXiv:1907.10326},
  year={2019}
}

@inproceedings{bhat2021adabins,
  title={Adabins: Depth estimation using adaptive bins},
  author={Bhat, Shariq Farooq and Alhashim, Ibraheem and Wonka, Peter},
  booktitle={Proceedings of the IEEE/CVF conference on computer vision and pattern recognition},
  pages={4009--4018},
  year={2021}
}

@inproceedings{patil2022p3depth,
  title={P3depth: Monocular depth estimation with a piecewise planarity prior},
  author={Patil, Vaishakh and Sakaridis, Christos and Liniger, Alexander and Van Gool, Luc},
  booktitle={Proceedings of the IEEE/CVF conference on computer vision and pattern recognition},
  pages={1610--1621},
  year={2022}
}

@inproceedings{ranftl2021vision,
  title={Vision transformers for dense prediction},
  author={Ranftl, Ren{\'e} and Bochkovskiy, Alexey and Koltun, Vladlen},
  booktitle={Proceedings of the IEEE/CVF international conference on computer vision},
  pages={12179--12188},
  year={2021}
}

@article{li2024binsformer,
  title={BinsFormer: Revisiting Adaptive Bins for Monocular Depth Estimation},
  author={Li, Zhenyu and Wang, Xuyang and Liu, Xianming and Jiang, Junjun},
  journal={IEEE Transactions on Image Processing},
  volume={33},
  pages={3964--3976},
  year={2024},
  publisher={IEEE}
}

@inproceedings{agarwal2023attention,
  title={Attention attention everywhere: Monocular depth prediction with skip attention},
  author={Agarwal, Ashutosh and Arora, Chetan},
  booktitle={Proceedings of the IEEE/CVF winter conference on applications of computer vision},
  pages={5861--5870},
  year={2023}
}

@inproceedings{bhat2022localbins,
  title={Localbins: Improving depth estimation by learning local distributions},
  author={Bhat, Shariq Farooq and Alhashim, Ibraheem and Wonka, Peter},
  booktitle={European Conference on Computer Vision},
  pages={480--496},
  year={2022},
  organization={Springer}
}

@article{kim2022global,
  title={Global-local path networks for monocular depth estimation with vertical cutdepth},
  author={Kim, Doyeon and Ka, Woonghyun and Ahn, Pyungwhan and Joo, Donggyu and Chun, Sehwan and Kim, Junmo},
  journal={arXiv preprint arXiv:2201.07436},
  year={2022}
}

@inproceedings{he2016deep,
  title={Deep residual learning for image recognition},
  author={He, Kaiming and Zhang, Xiangyu and Ren, Shaoqing and Sun, Jian},
  booktitle={Proceedings of the IEEE conference on computer vision and pattern recognition},
  pages={770--778},
  year={2016}
}

@inproceedings{liu2022convnet,
  title={A convnet for the 2020s},
  author={Liu, Zhuang and Mao, Hanzi and Wu, Chao-Yuan and Feichtenhofer, Christoph and Darrell, Trevor and Xie, Saining},
  booktitle={Proceedings of the IEEE/CVF conference on computer vision and pattern recognition},
  pages={11976--11986},
  year={2022}
}

@inproceedings{chenvision,
  title={Vision Transformer Adapter for Dense Predictions},
  author={Chen, Zhe and Duan, Yuchen and Wang, Wenhai and He, Junjun and Lu, Tong and Dai, Jifeng and Qiao, Yu},
  booktitle={The Eleventh International Conference on Learning Representations}
}

@inproceedings{lee2022mpvit,
  title={Mpvit: Multi-path vision transformer for dense prediction},
  author={Lee, Youngwan and Kim, Jonghee and Willette, Jeffrey and Hwang, Sung Ju},
  booktitle={Proceedings of the IEEE/CVF conference on computer vision and pattern recognition},
  pages={7287--7296},
  year={2022}
}

@article{wang2022pvt,
  title={Pvt v2: Improved baselines with pyramid vision transformer},
  author={Wang, Wenhai and Xie, Enze and Li, Xiang and Fan, Deng-Ping and Song, Kaitao and Liang, Ding and Lu, Tong and Luo, Ping and Shao, Ling},
  journal={Computational visual media},
  volume={8},
  number={3},
  pages={415--424},
  year={2022},
  publisher={TUP}
}

@inproceedings{huang2024localmamba,
  title={Localmamba: Visual state space model with windowed selective scan},
  author={Huang, Tao and Pei, Xiaohuan and You, Shan and Wang, Fei and Qian, Chen and Xu, Chang},
  booktitle={European conference on computer vision},
  pages={12--22},
  year={2024},
  organization={Springer}
}

@inproceedings{xiao2025spatial,
  title={SPATIAL-MAMBA: EFFECTIVE VISUAL STATE SPACE MODELS VIA STRUCTURE-AWARE STATE FUSION},
  author={Xiao, Chaodong and Li, Minghan and Zhang, Zhengqiang and Meng, Deyu and Zhang, Lei},
  booktitle={13th International Conference on Learning Representations, ICLR 2025},
  pages={44892--44910},
  year={2025},
  organization={International Conference on Learning Representations, ICLR}
}

@inproceedings{su2022vitas,
  title={Vitas: Vision transformer architecture search},
  author={Su, Xiu and You, Shan and Xie, Jiyang and Zheng, Mingkai and Wang, Fei and Qian, Chen and Zhang, Changshui and Wang, Xiaogang and Xu, Chang},
  booktitle={European Conference on Computer Vision},
  pages={139--157},
  year={2022},
  organization={Springer}
}

@inproceedings{liu2022uninet,
  title={Uninet: Unified architecture search with convolution, transformer, and mlp},
  author={Liu, Jihao and Huang, Xin and Song, Guanglu and Li, Hongsheng and Liu, Yu},
  booktitle={European Conference on computer vision},
  pages={33--49},
  year={2022},
  organization={Springer}
}

@inproceedings{li2024attnzero,
  title={Attnzero: efficient attention discovery for vision transformers},
  author={Li, Lujun and Wei, Zimian and Dong, Peijie and Luo, Wenhan and Xue, Wei and Liu, Qifeng and Guo, Yike},
  booktitle={European Conference on Computer Vision},
  pages={20--37},
  year={2024},
  organization={Springer}
}

@article{chen2021searching,
  title={Searching the search space of vision transformer},
  author={Chen, Minghao and Wu, Kan and Ni, Bolin and Peng, Houwen and Liu, Bei and Fu, Jianlong and Chao, Hongyang and Ling, Haibin},
  journal={Advances in Neural Information Processing Systems},
  volume={34},
  pages={8714--8726},
  year={2021}
}

@article{russakovsky2015imagenet,
  title={Imagenet large scale visual recognition challenge},
  author={Russakovsky, Olga and Deng, Jia and Su, Hao and Krause, Jonathan and Satheesh, Sanjeev and Ma, Sean and Huang, Zhiheng and Karpathy, Andrej and Khosla, Aditya and Bernstein, Michael and others},
  journal={International journal of computer vision},
  volume={115},
  number={3},
  pages={211--252},
  year={2015},
  publisher={Springer}
}

@inproceedings{yu2021pixelnerf,
  title={pixelnerf: Neural radiance fields from one or few images},
  author={Yu, Alex and Ye, Vickie and Tancik, Matthew and Kanazawa, Angjoo},
  booktitle={Proceedings of the IEEE/CVF conference on computer vision and pattern recognition},
  pages={4578--4587},
  year={2021}
}

@inproceedings{suhail2022generalizable,
  title={Generalizable patch-based neural rendering},
  author={Suhail, Mohammed and Esteves, Carlos and Sigal, Leonid and Makadia, Ameesh},
  booktitle={European Conference on Computer Vision},
  pages={156--174},
  year={2022},
  organization={Springer}
}

@inproceedings{du2023learning,
  title={Learning to render novel views from wide-baseline stereo pairs},
  author={Du, Yilun and Smith, Cameron and Tewari, Ayush and Sitzmann, Vincent},
  booktitle={Proceedings of the IEEE/CVF Conference on Computer Vision and Pattern Recognition},
  pages={4970--4980},
  year={2023}
}

@inproceedings{xu2024murf,
  title={Murf: Multi-baseline radiance fields},
  author={Xu, Haofei and Chen, Anpei and Chen, Yuedong and Sakaridis, Christos and Zhang, Yulun and Pollefeys, Marc and Geiger, Andreas and Yu, Fisher},
  booktitle={Proceedings of the IEEE/CVF Conference on Computer Vision and Pattern Recognition},
  pages={20041--20050},
  year={2024}
}

@inproceedings{charatan2024pixelsplat,
  title={pixelsplat: 3d gaussian splats from image pairs for scalable generalizable 3d reconstruction},
  author={Charatan, David and Li, Sizhe Lester and Tagliasacchi, Andrea and Sitzmann, Vincent},
  booktitle={Proceedings of the IEEE/CVF conference on computer vision and pattern recognition},
  pages={19457--19467},
  year={2024}
}

@inproceedings{hassani2023neighborhood,
  title={Neighborhood attention transformer},
  author={Hassani, Ali and Walton, Steven and Li, Jiachen and Li, Shen and Shi, Humphrey},
  booktitle={Proceedings of the IEEE/CVF conference on computer vision and pattern recognition},
  pages={6185--6194},
  year={2023}
}

@article{han2024demystify,
  title={Demystify mamba in vision: A linear attention perspective},
  author={Han, Dongchen and Wang, Ziyi and Xia, Zhuofan and Han, Yizeng and Pu, Yifan and Ge, Chunjiang and Song, Jun and Song, Shiji and Zheng, Bo and Huang, Gao},
  journal={Advances in neural information processing systems},
  volume={37},
  pages={127181--127203},
  year={2024}
}

@inproceedings{shaker2025groupmamba,
  title={GroupMamba: Efficient group-based visual state space model},
  author={Shaker, Abdelrahman and Wasim, Syed Talal and Khan, Salman and Gall, Juergen and Khan, Fahad Shahbaz},
  booktitle={Proceedings of the Computer Vision and Pattern Recognition Conference},
  pages={14912--14922},
  year={2025}
}

\vfill

\end{document}